\documentclass[journal]{IEEEtran}

\usepackage[utf8]{inputenc} % allow utf-8 input
\usepackage[T1]{fontenc}    % use 8-bit T1 fonts
\usepackage{hyperref}       % hyperlinks
\usepackage{url}            % simple URL typesetting
\usepackage{booktabs}       % professional-quality tables
\usepackage{amsfonts}       % blackboard math symbols
\usepackage{nicefrac}       % compact symbols for 1/2, etc.
\usepackage{microtype}      % microtypography
\usepackage{graphicx}       % added by xinrui
\usepackage{amsmath}        % added by xinrui
\usepackage{amssymb}  
\usepackage{bm}  
\usepackage{textcomp}
\usepackage{enumerate}
\usepackage{multirow}
\usepackage{diagbox}

\makeatletter  
\newif\if@restonecol  
\makeatother

\usepackage[linesnumbered,ruled,vlined]{algorithm2e}%[ruled,vlined]{  
\usepackage{algpseudocode}  
\usepackage{amsmath}  
\begin{document}

\title{CHAIN: Concept-harmonized Hierarchical Inference Interpretation of Deep Convolutional Neural Networks}

%\author{Dan~Wang, Xinrui~Cui,
%                 Septimiu E. Salcudean,~\IEEEmembership{Fellow,~IEEE}, and~Z.~Jane~Wang,~\IEEEmembership{Fellow,~IEEE}% <-this % stops a space
%\thanks{D. Wang, X. Cui, Septimiu E. Salcudean, and Z. J. Wang are with the Department
%of Electrical and Computer Engineering, University of British Columbia, BC, Canada. e-mail: (danw@ece.ubc.ca; xinruic@ece.ubc.ca; tims@ece.ubc.ca; zjanew@ece.ubc.ca.)}}
\author{Dan~Wang, Xinrui~Cui,
                 and~Z.~Jane~Wang,~\IEEEmembership{Fellow,~IEEE}% <-this % stops a space
\thanks{This work was supported in part by the Canadian Natural Sciences and Engineering Research Council (NSERC), the Four Year Doctoral Fellowship
and the International Doctoral Fellowship at the University of British Columbia.(Corresponding author: Xinrui Cui.)

D. Wang, X. Cui, and Z. J. Wang are with the Department
of Electrical and Computer Engineering, University of British Columbia, BC, Canada. e-mail: (danw@ece.ubc.ca; xinruic@ece.ubc.ca; zjanew@ece.ubc.ca.)}}
\maketitle

\begin{abstract}
With the great success of networks, it witnesses the increasing demand for the interpretation of the internal network mechanism, especially for the net decision-making logic.
%especially for how networks make their decisions for given inputs. With the widespread applications of deep convolutional neural networks (DCNNs), it becomes increasingly important for DCNNs not only to make accurate predictions but also to explain how they make their decisions.
To tackle the challenge, the Concept-harmonized HierArchical INference (CHAIN) is proposed to interpret the net decision-making process.
For net-decisions being interpreted, the proposed method presents the CHAIN interpretation in which the net decision can be hierarchically deduced into visual concepts from high to low semantic levels.
To achieve it, we propose three models sequentially, i.e., the concept harmonizing model, the hierarchical inference model, and the concept-harmonized hierarchical inference model.
Firstly, in the concept harmonizing model, visual concepts from high to low semantic-levels are aligned with net-units from deep to shallow layers.  
Secondly, in the hierarchical inference model, the concept in a deep layer is disassembled into units in shallow layers.
Finally, in the concept-harmonized hierarchical inference model, a deep-layer concept is inferred from its shallow-layer concepts.
After several rounds, the concept-harmonized hierarchical inference is conducted backward from the highest semantic level to the lowest semantic level.  
Finally, net decision-making is explained as a form of concept-harmonized hierarchical inference, which is comparable to human decision-making.
Meanwhile, the net layer structure for feature learning can be explained based on the hierarchical visual concepts. 
%Finally, net decision-making is explained as a form of concept-tree, which is comparable to human decision-making. 
In quantitative and qualitative experiments, we demonstrate the effectiveness of CHAIN at the instance and class levels.

\end{abstract}


\section{Introduction}
\label{sec:intro}
Recently, the convolution neural networks are applied to many different tasks, such as image classification \cite{xie2019self}, object detection \cite{pang2019libra}, and have achieved great success.
The network decision-making process is still a puzzle to us.
Therefore the interpretation models obtain more and more attention.
Among those, visual interpretation is a popular research direction owing to its similarity to the human understanding way.
The previous visual interpretations for internal features are focused on giving the pattern of the network on an individual layer \cite{olah2017feature}.
%%%

Notwithstanding, it does not provide an interpretable logical process for network decision-making.

The logic is interpretable in the decision-making process of human.
%There is a proposal that we can build an interpretation model in which the decision-making process of networks could be understood as a human-logic way. 
There is a question of whether the decision-making process of networks could be understood as a human-logic way. 
To answer it, we should solve the two following issues: 

\begin{enumerate}[$\bullet$]
%\item For visual tasks, we build our logic based on visual concepts, such as objects, parts, material, color.
%Do networks learn visual concepts in their feature representation?
\item For visual tasks, we build our logic based on visual concepts, i.e., color, material, part, object, scene.
%Visual concepts are from a low semantic level (color) to a high semantic level (scene).
The hierarchical structure of visual concepts starts from low to high semantic-levels.
In comparison, the stratified structure of networks builds from shallow to deep layers.
%Human-logic builds visual concepts from a low semantic level (color) to a high semantic level (scene). 
%For visual tasks, our logic is based on visual concepts, such as objects, parts, material, color.
Could the hierarchical structure for visual concepts be used to explain the stratified structure of networks in the visual task?
%Does the layer-structure of networks represent the stratified structure for visual concepts?
%%\item The hierarchical structure of visual concepts starts from a low semantic-level to a high semantic-level.
%In comparison, the hierarchical structure of networks starts from a shallow layer to a deep layer.
%What is the relationship between those two hierarchical structures?%Do networks build the hierarchical structure for visual concepts?
%%\item Human-logic builds visual concepts from a low semantic level (color) to a high semantic level (scene). Could the stratified structure for visual concepts be used to explain the layer-structure of networks?
\item In human-logic, a decision could be hierarchically deduced into sub-decisions.
Could the decision of networks be explained as the form of a hierarchical inference? 
%Could the decision of networks be explained as the form of a decision tree?
\end{enumerate}

\begin{figure*}[htp]
	\centering
	\includegraphics[width=1\linewidth]{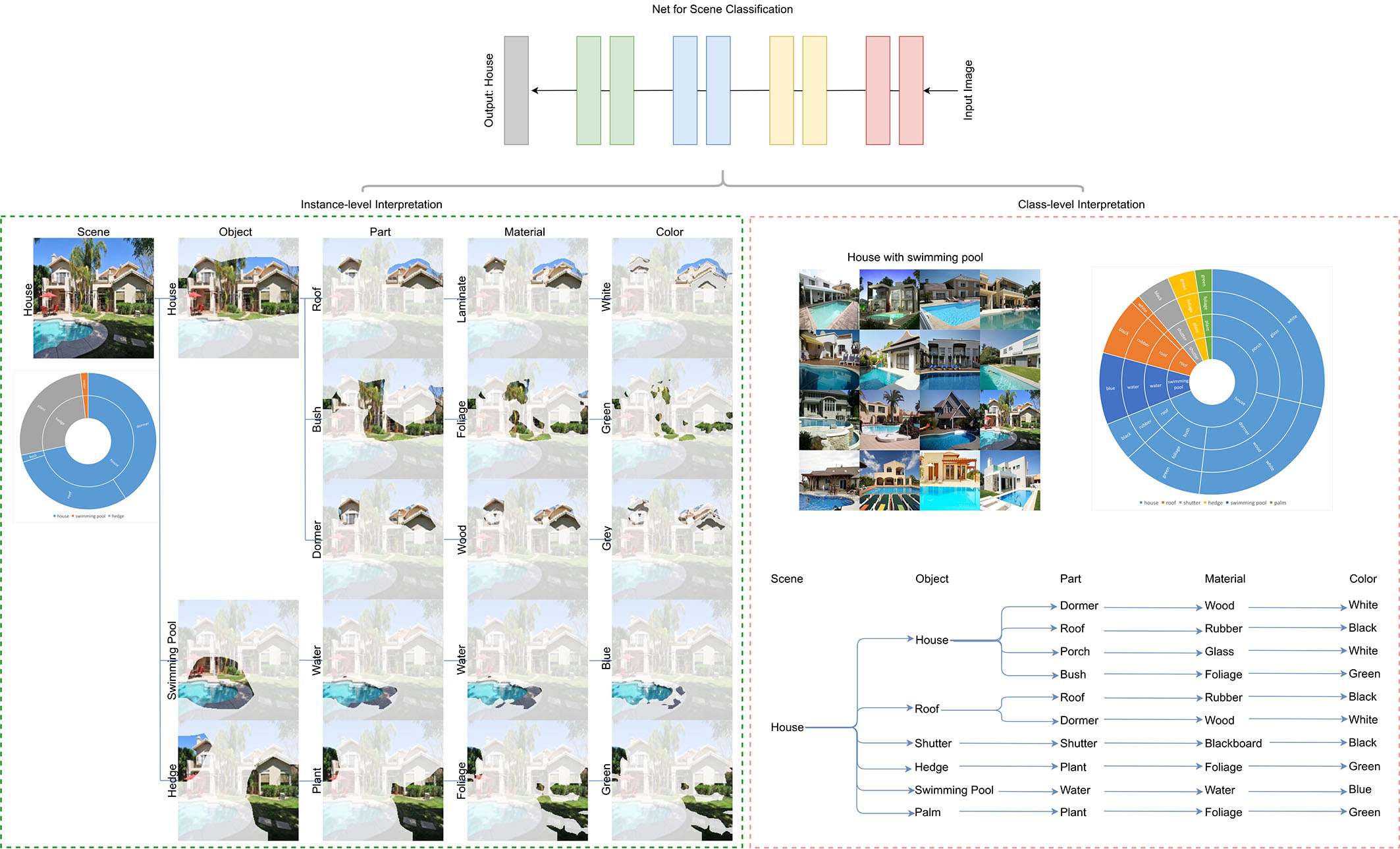}
	\centering
	\caption{Illustration of CHAIN interpretation. CHAIN can provide the instance-level and the class-level interpretation to explain the net decision-making process. It represents this hierarchical inference of net-units as an human-understandable explanation by utilizing different semantic-level visual concepts. }
	\label{fig:tree}
\end{figure*}

In this paper, we propose the framework of Concept HierArchical INference interpretation (CHAIN) which is inspired by the human understanding way.
Human logic is always designed from interpretable elements such as visual concepts for images.
Accordingly, for the first issue, we propose the concept harmonizing model which can interpret networks utilizing an interpretable visual concept.
In this model, network features from shallow to deep layers are harmonized with visual concepts from low to high semantic levels.

For us, high-level visual concepts such as scene can be deduced into low-level visual concepts such as object and part.
Therefore, for the second issue, we propose hierarchical inference model to decompose the concept in a deep layer into units in a shallow layer.
Subsequently, the concept-harmonized hierarchical inference model is introduced to infer a deep-layer concept into its shallow-layer concepts.
%In this model, based on the local linear regression, the deep-layer unit is represented as a sparse combination of shallow-layer units.
%Therefore, for the second issue, we propose hierarchical inference model to disassemble the concept-harmonized units from a deep layer to a shallow layer.
%In this model, based on the local linear regression, the deep-layer unit is represented as a sparse combination of shallow-layer units.
%the hierarchical inference model is proposed to hierarchically deduce the concept-harmonized unit of a deep layer into concept-harmonized units of a shallow layer.

Consequently, CHAIN explains a network decision by representing its concept-harmonized hierarchical inference in a human-understandable way.
%Subsequently, CHAIN explains a network decision by the concept-tree interpretation, which represents concept-harmonized hierarchical inference of networks in a human-understandable way.
The main contributions of this work are summarized as follows:
\begin{enumerate}[$\bullet$]
\item In the CHAIN interpretation, we explain net features in a stratified structure with visual concepts in a hierarchical structure.
%In the concept-tree, we interpret net features with visual concepts.
Specifically, we first build the concept harmonizing model in which visual concepts are aligned with net-units in a depth-stratified way (from deep to shallow layers).
%\item In the CHAIN interpretation, we explain net features with visual concepts.
%%In the concept-tree, we interpret net features with visual concepts.
%To achieve it, we build the concept harmonizing model in which visual concepts are aligned with net-units in a semantic-stratified way (from high to low semantic-levels). 
\item In the CHAIN interpretation, the feature learning of the network from shallow to deep layers is interpreted as a hierarchical logical process of the decision-making from low to high levels. 
%the layer structure of network learning is explained as a hierarchical logical structure of decision-making.
%In the concept-tree, the layer structure of network learning is explained as a hierarchical logical structure of decision-making.
To achieve that, we successively introduce the hierarchical inference model and the concept-harmonized hierarchical inference model.
Through the hierarchical inference model, the net learning for a deep-layer concept is inferred from shallow-layer features.
Based on the previous models, the concept-harmonized hierarchical inference model can hierarchically deduce a deep-layer concept into shallow-layer concepts.
%\item In the CHAIN interpretation, the layer structure of network learning is explained as a hierarchical logical structure of decision-making.
%%In the concept-tree, the layer structure of network learning is explained as a hierarchical logical structure of decision-making.
%Specifically, the hierarchical inference model is proposed to hierarchically deduce the concept-harmonized unit of a deep layer into concept-harmonized units of a shallow layer.
%\item The CHAIN can give the concept-harmonized hierarchical inference not only for a net decision of a specific instance but also for net decisions of a class. 
%The CHAIN can give the concept-tree interpretation not only for a net decision of a specific instance but also for net decisions of a class. 
\item For the instance level, CHAIN provides the concept-harmonized hierarchical inference of a net decision, which is an understandable logic for the network decision-making process from deep to shallow layers.
Moreover, for the class level, the CHAIN interpretation can explain net decisions for a class.
%The CHAIN provides the concept-tree interpretation, which is an understandable logic for the network decision-making process from deep to shallow layers.
\item In experiments, we analyze qualitatively and quantitatively the CHAIN interpretation for the intra- and inter-class.
\end{enumerate}

\begin{figure}[htp]
	\centering
	\includegraphics[width=1\linewidth]{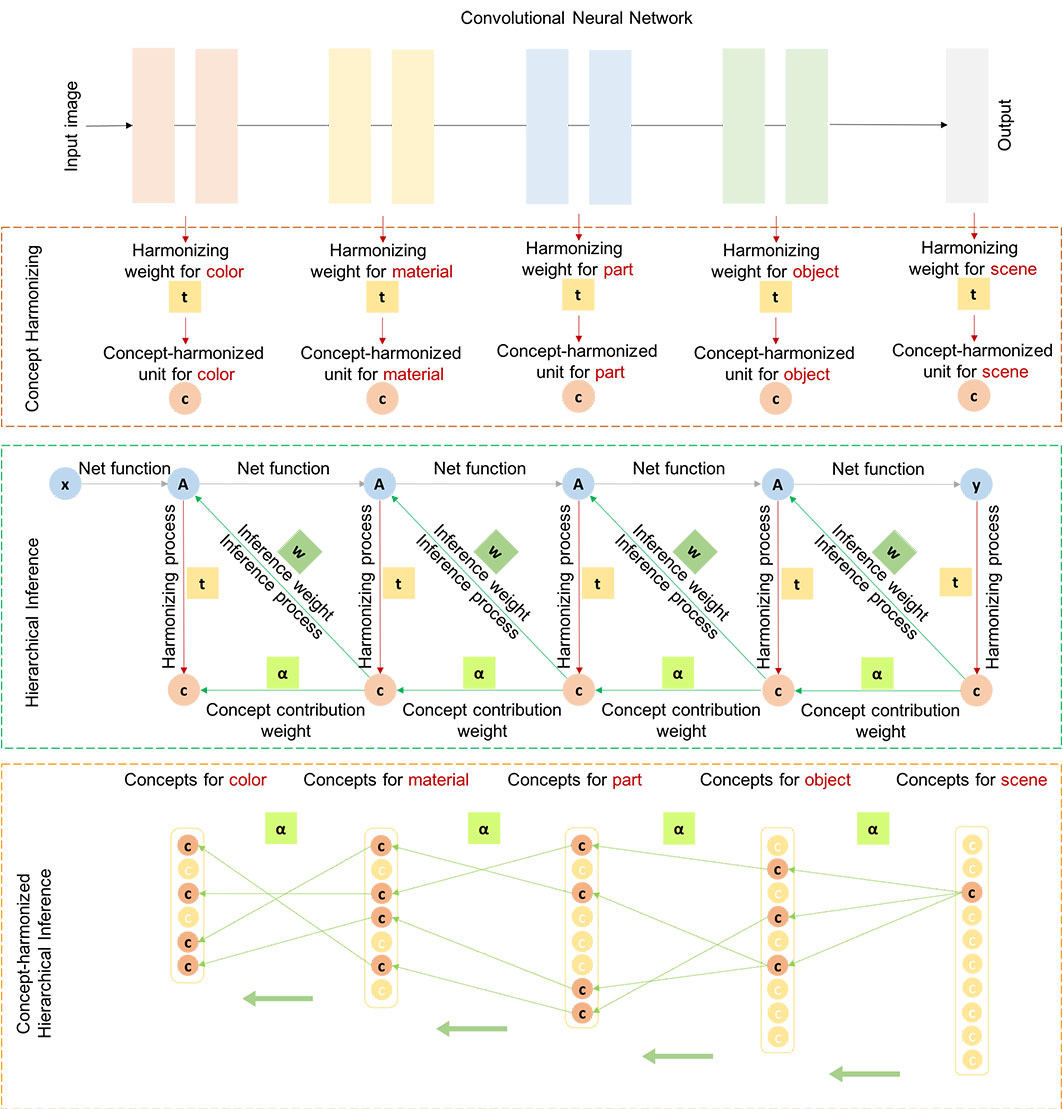}
	\centering
	\caption{The workflow of the CHAIN interpretation. In the concept harmonizing model, the units in the deep and shallow layers are matched with the high-level and low-level semantic visual concepts, respectively. In the hierarchical inference model, the deep-layer unit is represented as a sparse linear combination of shallow-layer units. The CHAIN interpretation further represents this hierarchical inference as an human-understandable explanation by utilizing different semantic-level visual concepts.}
	\label{fig:workflow}
\end{figure}

\section{Related Work} \label{relate}
Recently, the study of network interpretation has been increasingly drawn attention and gained popularity for it.
In this section, we review three main branches of network interpretation, as described below:
%There are a number of studies focusing on providing the visual interpretation for a particular network decision.
%It can be classified into two categories: one is based on perturbation methods, one is based on features in the last convolutional layer.

\textbf{Input-based network interpretation.}
The study in this direction explains network by learning critical input regions to a particular net output.
It utilizes the perturbation mechanism in which critical input regions are obtained by perturbing the input and observing output changes \cite{zeiler2014visualizing, Ribeiro2016, 8237633, 8653995}.
Therefore, it builds a mapping from the output space to the input space, which can give us interpretable visualization to understand net decisions.
Nonetheless, its visual interpretation is only based on input space which also means it can not explain the internal network mechanism.
Meantime, the shape of perturbation patches, which is a super-pixel  \cite{Ribeiro2016} or a regular grid \cite{zeiler2014visualizing}, restricts the shape of visual interpretation based on input images.
%For the perturbation methods, interpretation results are obtained to find importance regions by perturbing images.
%Therefore, their approaches are based on image-level visual.

\textbf{Feature-based network interpretation.}
Another popular interpretation technique is visualizing the internal features of networks.
By using the gradient information of net features, it can give the visual interpretation of internal networks.
Guided Backpropagation \cite{springenberg2014striving} and Deconvolution \cite{zeiler2014visualizing} visualize the image pattern which can obtain the largest activation of a particular net-unit.
However, it is only a general interpretation method which means it can not explain a specific net decision for a given input.
In comparison, CAM \cite{zhou2016learning} and Grad-CAM \cite{8237336} provide the class-discriminative interpretation of a specific net decision by visualizing the linear combination of features weighted by their gradients to the target output.
Nevertheless, those approaches heavily depend on the calculation of net gradient, which is not as efficient for shallow layers as that for the last layer.
To overcome it, CHIP model based on the channel-wise perturbation can distill class-discriminative channels from shallow to deep layers to interpret internal net-features \cite{8924894}.
Depending on the channel-wise perturbation, its performance is not limited by the perturbation-patch shape and the net-layer location.

\textbf{Concept-based network interpretation.}
To interpret net-features as human-understandable concepts, researchers proposed the concept-based interpretation \cite{pmlr-v80-kim18d}. 
The internal representation of networks can be interpreted with visual concepts by evaluating the overlap between a concept region and the saliency region of a net-feature \cite{8099837}.
However, it is a one-to-one alignment and not a learnable approach.
It means does not consider the one-to-many situation in which networks learn the representation of a visual concept from a combination of net-units.
Furthermore, Interpretable basis decomposition \cite{zhou2018interpretable} was proposed to represent the class-discriminative importance weight for the target class as a linear combination of concept-discriminative importance weights for different concepts.
Consequently, it can deduce the class-discriminative feature for the target class into concept-discriminative features for different concepts. 
It should be noticed that its decomposition target and bases need to utilize net features from the same layer in networks.
However, the study shows that different semantic-level concepts should be matched with net-features in different layers.
Therefore, in its interpretation for the final convolutional layer, low semantic-level concept bases might be as active as high semantic-level concept bases.
Meanwhile, this decomposition is limited to explain one internal layer in which the prediction is decomposed into the features in the last convolutional layer.
Therefore, it cannot explain the layer structure for the net decision-making process.

Here, we propose the CHAIN interpretation to explain the net decision-making from deep to shallow layers by the concept-harmonized hierarchical inference interpretation.
%Here, we propose the CHAIN to explain the net decision-making from deep to shallow layers by concept-tree interpretation.
Accordingly, layer-stratified net-features are explained by semantic-stratified visual concepts, which means net-features in different layers are aligned with visual concepts in different semantic-level.
The proposed interpretation is built by hierarchically inference of the concept from the deep to shallow layers.
%The concept-tree is built by hierarchically decomposing the concept-harmonized unit from the highest semantic-level layer to the lowest semantic-level layer.
Therefore, it can interpret hierarchical network learning as an interpretable decision-making process for visual concepts from high to low semantic levels.

\section{Methodology}
\label{sec:methodology}
\subsection{The Framework of Concept-harmonized Hierarchical Inference Interpretation}

For human, we have a knowledge system of visual-concepts from low to high semantic levels. 
According to it, the human decision-making process can be deduced into a series of sub-decision making processes from high to low levels.
In this paper, we propose CHAIN to builds a human-understandable interpretation.
CHAIN can explain the net decision-making process by the concept-harmonized hierarchical inference interpretation in which the operation of a network presents an analogy to the working of the human brain.

\begin{algorithm*} \label{algorithm}
\caption{Pseudocode of the Concept Harmonizing Algorithm}
\LinesNumbered 

\KwIn{the visual concepts in the $l$-th layer;\\
 \quad\quad\quad its network features in the $l$-th layer;\\
 \quad\quad\quad the dataset $\mathcal{D} \lbrace \mathbf{I}_n, \mathbf{a}^l_{n}, z^{l,k}_{n}\rbrace$ for the harmonizing model for the $k$-th concept in the $l$-th layer;} 
\KwOut{the optimal concept harmonizing weights in the target layer;} 
		  
		 \textbf{Initialization:} set $k = 1$, $\lambda > 0$; \\
		  \Repeat () {the concept harmonizing model is optimized for all concepts in the $l$-th layer}
		 {
		 \textbf{The optimization of the harmonizing model for the $k$-th concept in the $l$-th layer}:		 
		 $\arg\min_{\mathbf{t}^{l,k}} \quad  \frac{1}{2} \sum_{n=1}^{N} (\mathbf{t}^{l,k}\mathbf{a}^l_{n} - z^{l,k}_{n})^{2} + \lambda \|\mathbf{t}^{l,k}\|_{1} $ (in Eq. (\ref{concept harmonizing model}));\\

		 \textbf{return} the optimal harmonizing weight $\mathbf{t}^{l,k}$ of the $k$-th visual concept in the $l$-th layer;\\
		 %\textbf{return} the concept-harmonized unit $\mathbf{c}^{h,k}_{l}=\sum_{i} t^{h,k}_{l,i}\mathbf{A}^l_{i}$ in the $l$-th layer for the $k$-th concept in the $h$-th semantic level;\\ 
		 \textbf{update iteration:} $k \leftarrow k+1$;\\	 
		 }
		 \textbf{Return:} the optimal harmonizing weights of concepts in the target layer;\\
		  %\textbf{Return:} the concept-harmonized unit in the target semantic level;\\
\end{algorithm*} 

%%In this paper, in concept harmonizing model, visual concepts are linked to units in a similar semantic level.
%Specifically, the concept harmonizing model is proposed to harmonize net-units with visual concepts in a similar semantic level.
%Moreover, the hierarchical inference model interprets the network structure by disassembling the net-unit from deep to shallow layers.
%Consequently, in the concept-harmonized hierarchical inference model, the network can be explained as an interpretable decision-making process in which the hierarchical inference of net-units from deep to shallow layers is understood by utilizing visual concepts from high to low semantic-levels.
%%Based on the concept harmonizing and the hierarchical inference, the network can be explained as an interpretable concept-tree in which the hierarchical inference of net-units from deep to shallow layers is understood by utilizing visual concepts from high to low semantic-levels.

The CHAIN interpretation can be divided into three steps.
%In the first stage, net units in different layers will be harmonized with visual concepts in different semantic-levels.
%The semantic-stratified structure of visual concepts starts from a low semantic-level to a high semantic-level.
%In comparison, the layer-stratified structure of networks starts from a shallow layer to a deep layer.
In the first stage, we build a link between the semantic-stratified structure of visual concepts and the layer-stratified structure of networks.
Specifically, the concept harmonizing model is proposed to harmonize net-units with visual concepts in a similar semantic level.
%Specifically, net-units in the deep and shallow layers are harmonized with visual concepts in high and low semantic levels, respectively.
%%The Broden dataset is adopted as the visual concept dataset containing a hierarchical level of labeled visual-concept samples.
%%The visual concepts in Broden can be divided into five semantic-levels: the lowest-level visual concepts are colors and textures; the lower-level concept is the material-concept; the middle-level is part-concept; the higher-level is the object-concept; the highest-level is the scene-concept.
%In the harmonizing model, we can obtain the concept-harmonized unit for each visual concept from the highest semantic level to the lowest semantic level.
%%For scene recognition, the scene-concept corresponds to the output layer.
%%Moreover, the other four levels of visual concepts are matched with four selected layers from deep to shallow, respectively.
%% In each layer, the alignment between the labeled segmentation of corresponding visual concept and the activation map of the non-sparse unit is measured as the interpretability of the unit for the visual concept. 
Secondly, the hierarchical inference model interprets the network structure by disassembling the net-unit from deep to shallow layers.
Finally, in the concept-harmonized hierarchical inference model, the network can be explained as an interpretable decision-making process.
During this process, the high semantic-level concept in a deep layer is deduced into low semantic-level concepts in its shallow layer.
Based on that, the hierarchical inference of net-units from deep to shallow layers is understood by utilizing visual concepts from high to low semantic levels.
%the target concept in the deep layer is deduced into concepts in its shallow layer.
%The inference representation is learned by optimizing the hierarchical inference model.
%%a local linear regression problem.
%Subsequently, the contribution weight of the low semantic-level concept to the high semantic-level concept is learned from the concept-harmonized hierarchical inference model.
%%calculated according to the concept directional-derivative.
%Similarly, critical concepts in the shallow layer are then disassembled utilizing units in its shallower layer as sparse bases.
%After several rounds, the hierarchical inference proceeds from the deep layer to the shallow layer.
Consequently, CHAIN can present the optimal concept-harmonized hierarchical inference starting from deep to shallow layers.

In the CHAIN interpretation, the layer structure of networks is interpreted by the hierarchical structure of visual concepts.
Meanwhile, for a net prediction being interpreted, CHAIN can provide its interpretable hierarchical inference of the net decision-making.  

\subsection{Concept Harmonizing Model}
In the concept harmonizing stage, we design a model for learning the correlation between visual-concepts and net-units.
%For a specifc net layer, the visual-concepts with a similar semantic level are aligned with units in the target layer.
%For a specific semantic level, its visual-concepts are aligned with net-units with a similar semantic level.
Visual-concepts are aligned with net-units with a similar semantic level.

$\textbf{Visual concepts.}$
Here, we adopt the Broden dataset, which contains different semantic-level visual concepts.
Specifically, the visual concepts in concept harmonizing model have five semantic levels, i.e., color, material, part, object, and scene concepts from low to high level. 
Therefore, network features in five layers are selected to be harmonized with concepts in a similar semantic level. 
The samples of visual concepts are the pixel-level labeled concepts, excluding scene concept which cover full images.

$\textbf{The dataset of the concept harmonizing model.}$
For each concept, a concept harmonizing model is designed to learn its correlation with units in its corresponding semantic-level layer.
In the training of the harmonizing model for a particular concept, image pixels containing the target concept are positive samples. 
Otherwise, image pixels without the target concept are negative samples.
%For the positive sample, only the concept region is kept, and pixels in other regions are zero-out.
% except for that in the concept region.

The training samples for the target concept are fed into the network to obtain features in the corresponding layer.
Subsequently, we utilize the net sample-feature in the $l$-th layer denoted as $\mathbf{a}^l = [a^{l,1} ~ a^{l,2} \cdots a^{l,i} \cdots a^{l,I}]^T$ to learn the concept harmonizing model, where $a^{l,i}$ is the $i$-th net unit in the $l$-th layer for the target sample and $I$ is the number of units. 

Finally, for the harmonizing model of the $k$-th concept in the $l$-th layer, we get the dataset $\mathcal{D} \lbrace \mathbf{I}_n, \mathbf{a}^l_{n}, z^{l,k}_{n}\rbrace$ where for the $n$-th input sample $\mathbf{I}_n$, $\mathbf{a}^l_{n}$ is its $l$-th net-layer feature and $z^{l,k}_{n}$ is its concept label.

$\textbf{The harmonizing weights of units to the visual concept.}$
In the harmonizing model, units in the $l$-th layer are harmonized with the $k$-th concept in the corresponding semantic level by the harmonizing weight $\mathbf{t}^{l,k}= [t^{l,k}_{1} ~ t^{l,k}_{2} \cdots t^{l,k}_{i} \cdots t^{l,k}_{I}]$.
The harmonizing weight of the $i$-th unit is denoted as $t^{l,k}_{i}$.

\textbf{Concept harmonizing model.}
The harmonizing model of the $k$-th concept in the $l$-th layer is learned by optimizing the following problem:
\begin{equation} \label{concept harmonizing model}
\begin{split}
\arg\min_{\mathbf{t}^{l,k}} \quad & \frac{1}{2} \sum_{n=1}^{N} (\mathbf{t}^{l,k}\mathbf{a}^l_{n} - z^{l,k}_{n})^{2} + \lambda \|\mathbf{t}^{l,k}\|_{1}
\end{split}
\raisetag{2\normalbaselineskip}
\end{equation}
where $\lambda$ is the regularization parameter.
And the $z^{l,k}\in\{0,1\}$ is the label representing the absence (0) or presence (1) of the target concept.
%where $z^{h,k}=1$ indicates the presence of the concept,
%Finally, we can obtain the optimal correlation between net-units and visual-concepts in a similar semantic level.
After the optimization of Eq. (\ref{concept harmonizing model}), we can obtain the optimal correlation between the visual-concept and net-units in the corresponding layer.

\begin{algorithm*} \label{hierarchical inference}
\caption{Pseudocode of the Hierarchical Inference Algorithm}
\LinesNumbered 

\KwIn{for the image being interpreted, the deep-layer feature $\mathbf{A}^D$ and the shallow-layer feature $\mathbf{A}^S$;\\
% \quad\quad\quad the net function $\mathbf{A}^D_{j} = f_{\mathbf{A}^D_{j} }(  \mathbf{A}^S)$ mapping from shallow-layer features to deep-layer features;\\
\quad\quad\quad the optimal harmonizing weight $\mathbf{t}^{l,k}$ of the $k$-th visual concept in the deep layer;\\
%  \quad\quad\quad the target concept-harmonized unit $\mathbf{c}^{h,k}_{D}$ in the deep layer for the k-th concept in the h-th semantic-level;\\
%   \quad\quad\quad the concept-harmonized unit in the shallow layer $\mathbf{c}^{h-1,r}_{S}$;\\
%  \quad\quad\quad the net function $\mathbf{c}^{h,k}_{D}=f_{\mathbf{c}^{h,k}_{D} }(  \mathbf{A}^S)$ mapping from shallow-layer units to the concept-harmonized unit in the deep layer;\\
 \quad\quad\quad the perturbed dataset $\mathcal{D} \lbrace\mathbf{e}_n, \mathbf{x}_{n}, y^{D,k}_{n}\rbrace$ of the target concept;}  
 \KwOut{the optimal hierarchical inference for the target concept in the deep layer.}  
%\KwOut{the optimal hierarchical inference from concept-harmonized unit in the high semantic level to concept-harmonized units in the low semantic level;}  
%the optimal hierarchical inference from the concept-harmonized unit $\mathbf{c}^{h,k}_{D}$ in the deep layer to its shallow-layer units;
		 \textbf{Initialization:} set $i = 0$, $\mathbf{m}_{0}$, $\mathbf{u}_{0}$, $\lambda > 0$; \\
		
		  \textbf{The hierarchical inference optimization of the $k$-th visual concept in the deep layer:} %for the hierarchical inference of the target $q$-th deep-layer unit;\\
		 $\arg\min_{\mathbf{w}^{D,k}_{S}} \quad \frac{1}{2} \sum_{n=1}^{N} h(\mathbf{e}_{n}) (\mathbf{w}^{D,k}_{S}\mathbf{x}_{n} - y^{D,k}_{n})^{2} + \lambda \|\mathbf{w}^{D,k}_{S}\|_{1}$ (in Eq. (\ref{Hierarchical inference model}));\\
		 \Repeat () {stopping criterion is satisfied}
		 {		 			
		1:~
		$(\mathbf{w}^{D,k}_{S})_{i+1} \leftarrow (\sum_{n}h(\mathbf{e}_{n}) y^{D,k}_{n} {\mathbf{x}_{n}}^{T} + \rho \mathbf{m}_i + \rho \mathbf{u}_i)(\sum_{n}h(\mathbf{e}_{n}) \mathbf{x}_{n} {\mathbf{x}_{n}}^{T} + \rho \mathbf{I})^{-1}$;\\
		2:~ 
		 $\mathbf{m}_{i+1} \leftarrow soft ((\mathbf{w}^{D,k}_{S})_{i+1} - \mathbf{u}_i, \frac{\lambda}{\rho})$;\\
		3:~ update lagrange multipliers:\quad $\mathbf{u}_{i+1} \leftarrow \mathbf{u}_{i} - ((\mathbf{w}^{D,k}_{S})_{i+1} - \mathbf{m}_{i+1})$; \\
		4:~ \textbf{update iteration:} \quad $i \leftarrow i+1$;\\	

		 }
		 
		 \textbf{Return:} the optimal inference weight $\mathbf{w}^{D,k}_{S}$ of shallow-layer units to the target concept in the deep layer;\\
%                  \textbf{Return:} the contribution weight $s(\mathbf{c}^{h-1,r}_{S}, \mathbf{c}^{h,k}_{D})$ of a low semantic-level concept to the high semantic-level concept-harmonized unit;\\
		
\end{algorithm*}

Afterward, the optimization of the harmonizing model for the next critical concept in the target layer should be calculated, and so on and so forth.
Once this is done, we obtain the optimal concept harmonizing of net units in the same layer.
Subsequently, the optimization is conducted backward for the concept harmonizing of net units in the lower layer.
Finally, the concept harmonizing model can get the optimal concept harmonizing from the highest semantic level to the lowest semantic level.

\subsection{Hierarchical Inference Model}

In the previous stage, the concept harmonizing was designed to link units with concepts from high to low semantic levels.
%the deep-layer unit is disassembled into its importance shallow-layer units.
Based on it, in the hierarchical inference model, the concept in the deep layer can be inferred into the shallow layer. 
Therefore, a network decision is interpreted by representing its hierarchical inference.

\textbf{The inference weights of shallow-layer units to the concept in the deep layer.}
To explain the internal net structure, our CHAIN interpretation needs to optimize the importance of the shallow-layer units for a particular concept in its deep-layer.

For the $k$-th concept in the deep layer, the inference weight vector is $\mathbf{w}^{D,k}_{S} = [w^{D,k}_{S,1} ~ w^{D,k}_{S,2} \cdots w^{D,k}_{S,i} \cdots w^{D,k}_{S,I}]$, where $w^{D,k}_{S,i}$ represents the importance of the $i$-th unit in shallow layer to the target concept in the deep layer.
%And $I$ is the number of units in the shallow layer.

$\textbf{Net perturbation for hierarchical inference.}$
In CHAIN interpretation, we adopt the net perturbation-based approach in CHIP model.
The inference representation is learned by analyzing the variation of the concept in the deep layer after switching off its partial shallow-layer units. 
The underlying principle is that the concept in the deep layer would drop dramatically if the forward propagations of important shallow-layer units are blocked.

The pre-trained network is perturbed by shallow-layer gates to learn the inference weights. 
The shallow layer is associated with a gate layer in which each unit gate controls the state of the corresponding unit in the shallow layer.
%As shown in Fig. \ref{fig:illustration1}(top), the shallow layer is associated with a gate layer in which each unit gate controls the state of the corresponding unit in the shallow layer.
Here, a binary vector $\mathbf{e} = [e^1 ~ e^2 \cdots e^i \cdots e^{I}]^T$ is denoted the unit gate.
The $i$-th unit in the shallow layer is turned off if $e^i$ is zero.

The perturbed network is generated by adding the unit gate layer after the shallow layer.
We denote the original unit in the shallow layer as $\mathbf{A}^S \in \mathcal{R}^{U, V, I}$, where $U$ and $V$ are the width and height of the channel and $I$ is the number of units.
For the $i$-th unit in the shallow layer, the output of the shallow-layer gate layer is
\begin{equation}
\mathbf{\hat{A}}^S_{i} = e^i  \mathbf{A}^S_{i}
\end{equation}

The global average pooling of the shallow-layer gate is $\mathbf{x} = [x^1 ~ x^2 \cdots x^i \cdots x^{I}]^T$.
For the $i$-th shallow-layer unit, the global average pooling (GAP) is 
\begin{equation}
\begin{split}
x^i 
%= & \text{GAP}(\mathbf{\hat{A}}^S_p) 
= & \frac{1}{UV} \sum_{(u,v)} \mathbf{\hat{A}}^S_i
\end{split}
\end{equation}
%where $\text{GAP}(\mathbf{\hat{A}}^S_p) =  \frac{1}{UV} \sum_{(u,v)} \mathbf{\hat{A}}^S_p$ for the shallow layer.

\textbf{The concept-harmonized unit.}
Based on the optimal harmonizing weight, the concept-harmonized unit $\mathbf{c}^{D,k}$ for the $k$-th concept in the deep layer is defined as
\begin{equation} 
\begin{split}
\mathbf{c}^{D,k}=\sum_{j} t^{D,k}_{j}\mathbf{A}^D_{j}
\end{split}
\raisetag{2\normalbaselineskip}
\end{equation}
where $\mathbf{A}^D_{j}$ means the $j$-th feature in the deep layer.
%$t^{l,k}_{i}$ is the optimal harmonizing weight of the $i$-th units in the $l$-th layer to the $k$-th concept in the corresponding semantic level.

The net function mapping from shallow-layer features to deep-layer features is denoted as $f_{\mathbf{A}}(\cdot)$.
Specifically, the original feature of the $j$-th deep-layer unit is expressed as 
\begin{equation}
\mathbf{A}^D_{j} = f_{\mathbf{A}^D_{j} }(  \mathbf{A}^S)
\end{equation}
%and its global average pooling is recorded as
%\begin{equation}
%a^q = \frac{1}{UV} \sum_{(u,v)} \mathbf{A}^D^q
%\end{equation}

In the perturbed network, we add control gate layers behind the shallow layer without changing the original weights from the shallow to the deep layer in the pretrained network. 
%We can obtain different perturbed networks by changing the values in control gate layers.

%The net function mapping from shallow-layer units to the concept in the deep layer is denoted as $f_{\mathbf{c}}(\cdot)$.
%Specifically, the concept-harmonized unit $\mathbf{c}^{D,k}$ for the $k$-th concept in the deep layer is defined as
%\begin{equation} 
%\begin{split}
%\mathbf{c}^{D,k}=&f_{\mathbf{c}^{D,k}}(  \mathbf{A}^S)\\
%=&\sum_{j} (t^{D,k}_{j} \cdot \mathbf{A}^D_{j})\\
%=& \sum_{j} (t^{D,k}_{j} \cdot f_{\mathbf{A}^D_{j} }(  \mathbf{A}^S) )
%\end{split}
%\raisetag{2\normalbaselineskip}
%\end{equation}
%where $\mathbf{t}^{D,k}_{j}$ is the optimal harmonizing weight of the $j$-th units in the deep layer to the $k$-th concept in the $h$-th semantic level.

After the net perturbation, the concept-harmonized unit in the deep layer mapping from perturbed shallow-layer features is
\begin{equation}
\mathbf{\hat{c}}^{D,k}= \sum_{j} (t^{D,k}_{j} \cdot f_{\mathbf{A}^D_{j} }(  \mathbf{\hat{A}}^S) )
%= &f_{\mathbf{c}^{h,k}_{D} }(  \mathbf{\hat{A}}^S)
\end{equation}

In perturbed net, its global average pooling is
\begin{equation}
y^{D,k}= \frac{1}{U^{\prime}V^{\prime}} \sum_{(u',v')} \mathbf{\hat{c}}^{D,k}
\end{equation}
%========
$\textbf{The dataset of the hierarchical inference model.}$ 
In order to learn the inference weights of shallow-layer units, we need to generate the perturbed dataset. 
Specifically, to learn $\mathbf{w}^{D,k}_{S}$, the perturbed dataset is obtained by the following three steps: 
%, the inference weight vector of the $k$-th concept in the deep layer
The first step is to generate the perturbed networks. For the shallow layer, we sample the channel gate values by using $\mathcal{D}_{gate} = {\{ \mathbf{e}_n \}}_{n=1}^N$. And for other layers, we freeze the channel gate to be open.

Secondly, we feed each image into each perturbed network and get the features of shallow and deep layers. 
In $n$-th perturbed network based on $\mathbf{e}_n$, the global average pooling of the shallow layer is denoted as $\mathbf{x}_{n}$.
%the global average pooling of the $i$-th shallow-layer unit is 
%\begin{equation}
%\begin{split}
%x^i_{n} =& \frac{1}{UV} \sum_{(u,v)} e^i_n  \mathbf{A}^S_{i} \\
%%=& \frac{1}{UV} \sum_{(u,v)} \mathbf{\hat{A}}^S_i( e^i_n) \\
%\end{split}
%\end{equation}
%and the global average pooling of the shallow layer is denoted as $\mathbf{x}_{n}= [x^1_{n} ~ x^2_{n} \cdots x^i_{n} \cdots x^{I}_{n}]^T$.

Likewise, in the $n$-th perturbed network, the global average pooling of the $k$-th concept in the deep layer is denoted as $y^{D,k}_{n}$.
%the average value of the concept-harmonized unit in the deep layer for the $k$-th concept in the $h$-th semantic level is denoted as $y^{h,k}_{n}$.
%\begin{equation}
%y^{h,k}_{n} = \frac{1}{U^{\prime}V^{\prime}} \sum_{(u,v)} \mathbf{\hat{c}}^{h,k}_{D}(\mathbf{e}_n) 
%\end{equation}

%Meanwhile, for the image, its global average pooling of the $q$-th unit in the original deep-layer is recorded as 
%\begin{equation}
%a^q_m= \frac{1}{UV} \sum_{(u,v)} \mathbf{A}^D_q(\mathbf{I}_m) 
%\end{equation}
%Here, $a^q_m$ can also measure the feature activation of image $\mathbf{I}_m$ to the $q$-th deep-layer unit.

Finally, for the $k$-th concept in the deep layer of the image being interpreted, we get the perturbed dataset $\mathcal{D} \lbrace\mathbf{e}_n, \mathbf{x}_{n}, y^{D,k}_{n}\rbrace$.
% of the target concept-harmonized unit $\mathbf{\hat{c}}^{h,k}_{D}$.

\textbf{Hierarchical inference model.}
The inference representation is optimized by solving a local linear regression problem on a net-perturbation dataset.
Given the perturbed dataset $\mathcal{D}$ of the $k$-th concept in the deep layer, we formulate the hierarchical inference model as 
\begin{equation} \label{Hierarchical inference model} 
\begin{split}
\arg\min_{\mathbf{w}^{D,k}_{S}} \quad & \frac{1}{2} \sum_{n=1}^{N} h(\mathbf{e}_{n}) (\mathbf{w}^{D,k}_{S}\mathbf{x}_{n} - y^{D,k}_{n})^{2} + \lambda \|\mathbf{w}^{D,k}_{S}\|_{1}
\end{split}
\raisetag{2\normalbaselineskip}
\end{equation}
where $\lambda$ is the regularization parameter.

In CHAIN model, the first term in our interpretation model is the loss function.
In the loss function, $h(\mathbf{e}_{n})$ is denoted as the proximity measure between a binary channel gate vector $\mathbf{e}$ and the all-one vector $\mathbf{1}$. 
Specifically, it is defined as
\begin{equation}
h(\mathbf{e}_{n}) = \exp(-\frac{1}{\sigma^{2}}\lVert \mathbf{e}_{n} - \mathbf{1} \rVert_{2}^{2})
\end{equation}

The second term is the sparse regularization term owing to the inherent sparse property of network structure. 
Meanwhile, to make the interpretation model be simple enough to be interpretable, the sparsity of inference weights measures the complexity of the interpretation model. 

%The optimization problem in Eq. (\ref{interpret1-2}) is convex and can be solved based on ADMM \cite{afonso2011augmented}.
Here, to solve the optimization problem in Eq. (\ref{Hierarchical inference model}), we design a hierarchical inference algorithm by adopting the alternating iteration rule to learn $\mathbf{w}^{D,k}_{S}$.

\textbf{Hierarchical inference algorithm.}
The optimization problem can be converted into the equivalent formulation
\begin{equation}
\begin{split}
\arg\min_{\mathbf{w}^{D,k}_{S},\mathbf{m}}& \frac{1}{2} \sum_{n=1}^{N} h(\mathbf{e}_{n}) (\mathbf{w}^{D,k}_{S}\mathbf{x}_{n} - y^{D,k}_{n})^{2} + \lambda \|\mathbf{m}\|_{1} \\
\text{subject to}& \quad \mathbf{w}^{D,k}_{S} = \mathbf{m}
\end{split}
\raisetag{2\normalbaselineskip}
\end{equation}

The augmented Lagrangian for the above problem is
\begin{equation}
\begin{split}
%\arg\min_{\mathbf{w}^{D,k}_{S}, \mathbf{m}, \mathbf{v}} &\frac{1}{2} \sum_{n=1}^{N} h(\mathbf{e}_{n}) (\mathbf{w}^{D,k}_{S}\mathbf{x}_{n} - y^{D,k}_{n})^{2} + \lambda \|\mathbf{m}\|_{1} + \mathbf{v}^T (\mathbf{w}^{D,k}_{S} - \mathbf{m}) + \frac{\rho}{2} \Vert \mathbf{w}^{D,k}_{S} - \mathbf{m} \Vert_2^2
\arg\min_{\mathbf{w}^{D,k}_{S}, \mathbf{m}, \mathbf{v}} \quad &\frac{1}{2} \sum_{n=1}^{N} h(\mathbf{e}_{n}) (\mathbf{w}^{D,k}_{S}\mathbf{x}_{n} - y^{D,k}_{n})^{2} + \lambda \|\mathbf{m}\|_{1}\\
& + \mathbf{v}^T (\mathbf{w}^{D,k}_{S} - \mathbf{m}) + \frac{\rho}{2} \Vert \mathbf{w}^{D,k}_{S} - \mathbf{m} \Vert_2^2
\end{split}
\raisetag{2\normalbaselineskip}
\end{equation}

\begin{algorithm*} \label{Concept-harmonized hierarchical inference}
\caption{Pseudocode of the Concept-harmonized Hierarchical Inference Algorithm}
\LinesNumbered 

\KwIn{for the image being interpreted, the optimal inference weight $\mathbf{w}^{D,k}_{S}$ of shallow-layer units for the k-th concept in the deep layer;\\
 \quad\quad\quad the harmonizing weight set $ \mathbf{\Phi}_S$ of net units for concepts in the shallow layer; }
 
\KwOut{the optimal concept-harmonized hierarchical inference from the k-th concept in the deep layer to concepts in the shallow layer;}  
%the optimal hierarchical inference from the concept-harmonized unit $\mathbf{c}^{h,k}_{D}$ in the deep layer to its shallow-layer units
		
		  \textbf{The concept-harmonized hierarchical inference optimization for the k-th concept in the deep layer:} %for the hierarchical inference of the target $q$-th deep-layer unit;\\
		 $\arg\min_{\mathbf{\alpha}^{D,k}_{S}} \quad    \| \mathbf{\Phi}_{S}\mathbf{\alpha}^{D,k}_{S} - (\mathbf{w}^{D,k}_{S})^T\|_{2}^{2} \quad \textup{subject to}  \quad \|\mathbf{\alpha}^{D,k}_{S}\|_{0}  \leqslant \epsilon$ (in Eq. (\ref{Concept-harmonized hierarchical inference model}));\\
		 
                  \textbf{Return:} the contribution weight vector  $ \mathbf{\alpha}^{D,k}_{S}$ of concepts in the shallow layer to the k-th concept in the deep layer;\\
		
\end{algorithm*} 

The equation can be rewritten as 
\begin{equation}
\begin{split}
%\arg\min_{\mathbf{w}^{D,k}_{S}, \mathbf{m}, \mathbf{u}} &\frac{1}{2} \sum_{n=1}^{N} h(\mathbf{e}_{n}) (\mathbf{w}^{D,k}_{S}\mathbf{x}_{n} - y^{D,k}_{n})^{2} + \lambda \|\mathbf{m}\|_{1} + \frac{\rho}{2} \Vert \mathbf{w}^{D,k}_{S} - \mathbf{m} - \mathbf{u} \Vert_2^2
\arg\min_{\mathbf{w}^{D,k}_{S}, \mathbf{m}, \mathbf{u}} &\frac{1}{2} \sum_{n=1}^{N} h(\mathbf{e}_{n}) (\mathbf{w}^{D,k}_{S}\mathbf{x}_{n} - y^{D,k}_{n})^{2}+ \lambda \|\mathbf{m}\|_{1}\\
& + \frac{\rho}{2} \Vert \mathbf{w}^{D,k}_{S} - \mathbf{m} - \mathbf{u} \Vert_2^2
\end{split}
\raisetag{2\normalbaselineskip}
\end{equation}
where 
\begin{equation}
\mathbf{u} \equiv - \frac{1}{\rho} \mathbf{v}
\end{equation}

Through a careful choice of the new variable, the initial problem is converted into a simple problem.
Given that the optimization is considered over the variable $\mathbf{w}^{D,k}_{S}$, the optimization function can be reduced to
\begin{equation}
\begin{split}
%\mathbf{w}^{D,k}_{S} \leftarrow \arg\min_{\mathbf{w}^{D,k}_{S}} &\frac{1}{2} \sum_{n=1}^{N} h(\mathbf{e}_{n}) (\mathbf{w}^{D,k}_{S}\mathbf{x}_{n} - y^{D,k}_{n})^{2} + \frac{\rho}{2} \Vert \mathbf{w}^{D,k}_{S} - \mathbf{m} - \mathbf{u} \Vert_2^2
\mathbf{w}^{D,k}_{S} \leftarrow \arg\min_{\mathbf{w}^{D,k}_{S}} &\frac{1}{2} \sum_{n=1}^{N} h(\mathbf{e}_{n}) (\mathbf{w}^{D,k}_{S}\mathbf{x}_{n} - y^{D,k}_{n})^{2}\\
&+ \frac{\rho}{2} \Vert \mathbf{w}^{D,k}_{S} - \mathbf{m} - \mathbf{u} \Vert_2^2
\end{split}
\raisetag{2\normalbaselineskip}
\end{equation}

The solution is
\begin{equation}
\begin{split}
%(\mathbf{w}^{D,k}_{S})_{i+1} \leftarrow &(\sum_{s,n}h(\mathbf{e}_{n}) y^{D,k}_{n} {\mathbf{x}_{n}}^{T} + \rho \mathbf{m}_i + \rho \mathbf{u}_i)(\sum_{s,n}h(\mathbf{e}_{n}) \mathbf{x}_{n} {\mathbf{x}_{n}}^{T} + \rho \mathbf{I})^{-1}
(\mathbf{w}^{D,k}_{S})_{i+1} \leftarrow &(\sum_{n}h(\mathbf{e}_{n}) y^{D,k}_{n} {\mathbf{x}_{n}}^{T} + \rho \mathbf{m}_i \\
&+ \rho \mathbf{u}_i)(\sum_{n}h(\mathbf{e}_{n}) \mathbf{x}_{n} {\mathbf{x}_{n}}^{T} + \rho \mathbf{I})^{-1}
\end{split}
\raisetag{2\normalbaselineskip}
\end{equation}

In order to calculate $\mathbf{m}$, the optimization problem to be solved is
\begin{equation}
\mathbf{m} \leftarrow \arg \min_{\mathbf{m}} \lambda \Vert \mathbf{m} \Vert_1 + \frac{\rho}{2} \Vert \mathbf{w}^{D,k}_{S} - \mathbf{m} - \mathbf{u} \Vert_2^2
\end{equation}

The solution is
\begin{equation}
\mathbf{m}_{i+1} \leftarrow soft ((\mathbf{w}^{D,k}_{S})_{i+1} - \mathbf{u}_i, \frac{\lambda}{\rho})
\end{equation}

Lagrange multipliers update to
\begin{equation}
\mathbf{u}_{i+1} \leftarrow \mathbf{u}_{i} - ((\mathbf{w}^{D,k}_{S})_{i+1} - \mathbf{m}_{i+1})
\end{equation}

By the optimization of Eq. (\ref{Hierarchical inference model}), we obtain the optimal inference weight of shallow layer units to the $k$-th concept in the deep layer.

\subsection{Concept-harmonized Hierarchical Inference Model}

In previous stages, we complete the concept harmonizing and the hierarchical inference separately.
%the deep-layer unit is disassembled into its importance shallow-layer units.
By combining them, a network decision can be interpreted by representing its concept-harmonized hierarchical inference.
%Specifically, utilizing the concept directional-derivative, we study the contribution of low semantic-level concepts by shallow-layer units to the high semantic-level concept-harmonized unit in the deep layer.
%It can also be regarded as the inference of the concept-harmonized unit in the high semantic level into concept-harmonized units in the low semantic level.

\textbf{Concept-harmonized hierarchical inference model.}
Specifically, we deduce the concept in the high semantic level into concepts in the low semantic level.
Based on it, the contributions of low-level concepts in shallow-layer to a high-level concept in the deep-layer are computed.

The contribution weight vector of concepts in the shallow layer to the k-th concept in the deep layer is defined as $ \mathbf{\alpha}^{D,k}_{S}= [\alpha^{D,k} _{S,1}~\alpha^{D,k}_{S,2} \cdots \alpha^{D,k}_{S,k^{\prime}} \cdots \alpha^{D,k}_{S,K_{S}}]^T$.
Specifically, $\alpha^{D,k}_{S,k^{\prime}}$ is the contribution of the $k^{\prime}$-th concept in the shallow layer to the k-th concept in the deep layer. 
The harmonizing set of concepts in the shallow-layer is denoted as $ \mathbf{\Phi}_S= [\mathbf{t}^{S, 1}; \mathbf{t}^{S, 2}; \cdots ; \mathbf{t}^{S, K_{S}} ]^T$.
$\mathbf{t}^{S,k^{\prime}}$ denotes the harmonizing weight of the $k^{\prime}$-th concept in the shallow layer. % $\mathbf{t}^{r}_{S} \in \mathcal{R}^{1, I_{S} }$
And the inference weight $\mathbf{w}^{D, k}_{S}$ is the importance of shallow-layer units to the k-th concept in the deep layer.% $\mathbf{w}^{D, k}_{S} \in \mathcal{R}^{1, I_{S} }$

%When the $K_{S} > I_{S}$
The concept-harmonized hierarchical inference model is formulated as

\begin{equation} \label{Concept-harmonized hierarchical inference model}
\begin{split}
\arg\min_{\mathbf{\alpha}^{D,k}_{S}} \quad &   \| \mathbf{\Phi}_{S}\mathbf{\alpha}^{D,k}_{S} - (\mathbf{w}^{D,k}_{S})^T\|_{2}^{2} \\
\textup{subject to}  \quad &\|\mathbf{\alpha}^{D,k}_{S}\|_{0}  \leqslant \epsilon
\end{split}
\raisetag{2\normalbaselineskip}
\end{equation}
where $\|\cdot\|_{0}$ refers to the number of nonzero elements in the vector and is also viewed as the measure of sparsity. 
Moreover, the concept-harmonized hierarchical inference sparsity is bounded by $\epsilon$.

Based on the contribution weight, the high level concept is inferred into low level concepts.
% through deep- to shallow- layer units.
Meanwhile, we can give a quantitative analysis of the concept-harmonized hierarchical inference interpretation for a net decision.

%===
Then, the concept-harmonized hierarchical inference models of the other critical concepts in the deep layer are continuously optimized.
%And the critical low level concepts in the shallow layer to its high level concepts in the deep layer are selected as the next targets to be deduced. 
Subsequently, the optimization is conducted backward for the concept-harmonized hierarchical inference model from the deep to the shallow layer.
Finally, the concept-harmonized hierarchical inference model can get the optimal hierarchical inference representation of concepts from the highest to the lowest semantic level.

%\begin{equation} 
%\begin{split}
%\arg\min_{\mathbf{\alpha}^{h,k}_{h-1}} \quad &   \|\mathbf{\alpha}^{h,k}_{h-1}\|_{0}  \\
%\textup{subject to}  \quad &\| \mathbf{T}^{h-1}_{S}\mathbf{\alpha}^{h,k}_{h-1} - \mathbf{w}^{h,k}_{S}\|_{2}^{2} \leqslant \epsilon
%\end{split}
%\raisetag{2\normalbaselineskip}
%\end{equation}
%where the concept-harmonized hierarchical inference noise is bounded by $\epsilon$.
%
%\begin{equation} 
%\begin{split}
%\arg\min_{\mathbf{\alpha}^{h,k}_{h-1}} \quad & \frac{1}{2} \|\mathbf{T}^{h-1}_{S}\mathbf{\alpha}^{h,k}_{h-1} - \mathbf{w}^{h,k}_{S}\|_{2}^{2} + \lambda \|\mathbf{\alpha}^{h,k}_{h-1}\|_{1}
%\end{split}
%\raisetag{2\normalbaselineskip}
%\end{equation}
%where $\lambda$ is the regularization parameter.

\begin{figure*}[htp]
	\centering
	\includegraphics[width=1\linewidth]{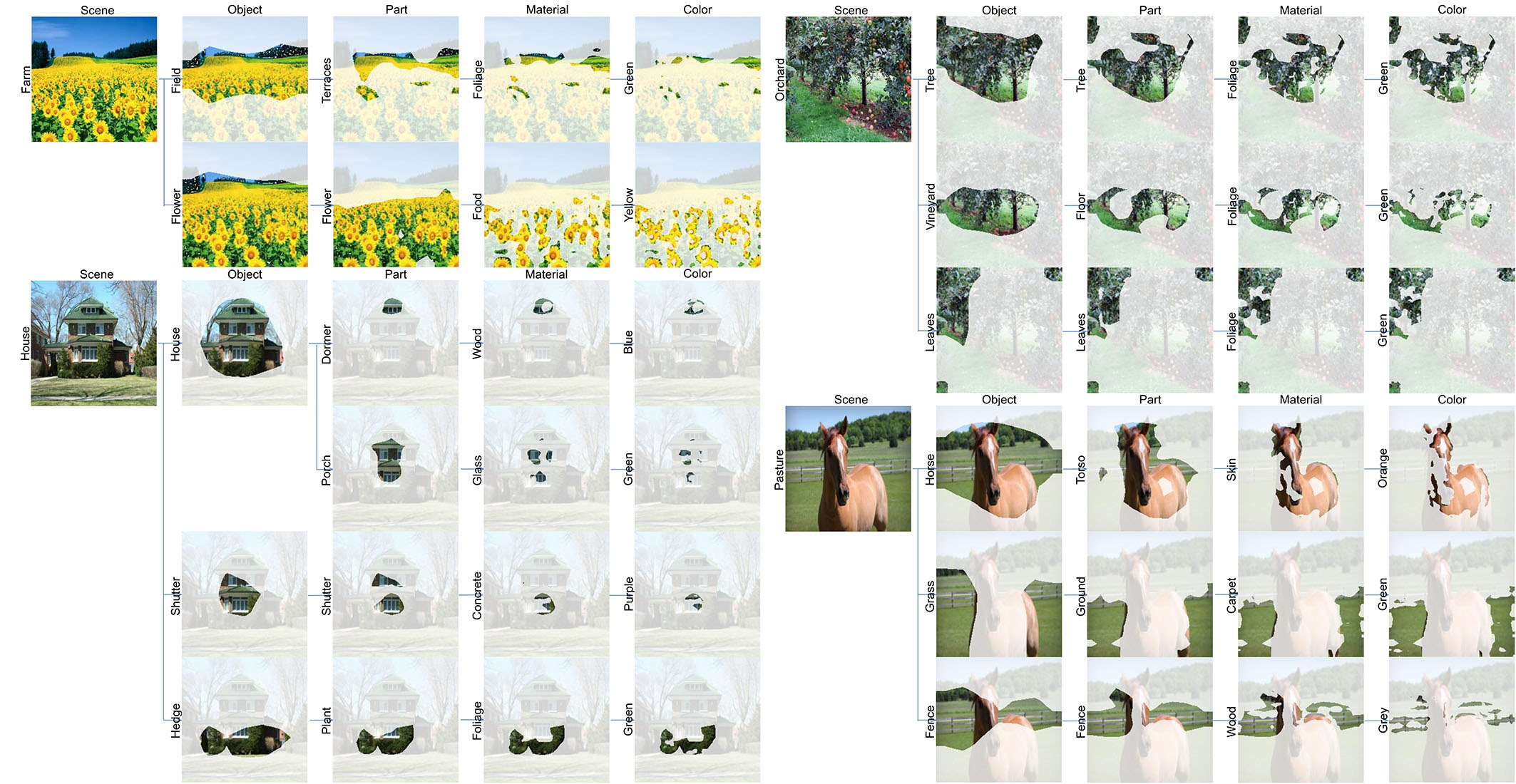}
	\centering
	\caption{Instance-level CHAIN interpretation for images from four classes (i.e. Farm, Orchard, House, and Pasture).}
	\label{fig:instance-inter}
\end{figure*}

$\textbf{The concept directional-derivative.}$
For the inference of part concepts, we only care about the most activated material concept which has the most significant contribution to the target part concept. 
It is also applied to the inference from the material concept to the color concept.
Therefore, we design a simple way to select the most critical shallow-layer concept to the target deep-layer concept.

Specifically, utilizing the concept directional-derivative, we study the contribution of shallow-layer concepts to the deep-layer concept.
The concept directional-derivative is defined as
\begin{equation} 
\begin{split}
\nabla_{\mathbf{t}^{S,k^{\prime}}} f_{\mathbf{c}^{D,k}}(  \mathbf{A}^S) & = \lim_{\delta \rightarrow 0}\frac{f_{\mathbf{c}^{D,k}}(  \mathbf{A}^S+\delta\mathbf{t}^{S,k^{\prime}})-f_{\mathbf{c}^{D,k}}(  \mathbf{A}^S)}{\delta} \\
%= &\nabla g^q(  \mathbf{A}^S) \cdot \mathbf{t}^k_S\\
&= \frac{\partial f_{\mathbf{c}^{D,k}}(  \mathbf{A}^S) }{\partial \mathbf{A}^S} \cdot \mathbf{t}^{S,k^{\prime}}\\
\end{split}
\raisetag{2\normalbaselineskip}
\end{equation}
which is the directional derivative of the deep-layer concept function $f_{\mathbf{c}^{D,k}}(\cdot)$ along the concept direction $ \mathbf{t}^{S,k^{\prime}}$ at the $\mathbf{A}^S$ in the shallow-layer feature space.
In the concept harmonizing model, $ \mathbf{t}^{S,k^{\prime}}$ is also the direction of the $k^{\prime}$-th concept at the shallow-layer feature space.
%the harmonizing weight of units in the shallow-layer to the $k^{\prime}$-th concept in the $(h-1)$-th semantic level.
%It is also the direction of the low semantic-level concept at the shallow-layer feature space.
In mathematics, the concept directional-derivative represents the instantaneous rate of change of the function $ f_{\mathbf{c}^{D,k}}(\cdot)$, moving through $\mathbf{A}^S$ with a velocity specified by $ \mathbf{t}^{S,k^{\prime}}$.

In the hierarchical inference model, it is assumed that $\frac{\partial f_{\mathbf{c}^{D,k}}(  \mathbf{A}^S) }{\partial \mathbf{A}^S} \approx  \mathbf{w}^{D,k}_{S}$. 
Therefore, the concept directional-derivative can be rewritten as 
\begin{equation} 
\begin{split}
\nabla_{\mathbf{t}^{S,k^{\prime}}} f_{\mathbf{c}^{D,k}}(  \mathbf{A}^S) & \approx  \mathbf{w}^{D,k}_{S} \cdot \mathbf{t}^{S,k^{\prime}}\\
%&= \sum_{i=1} w^{h,k}_{S,i} \cdot t^{h-1,k^{\prime}}_{S,i}
\end{split}
\raisetag{2\normalbaselineskip}
\end{equation}
which is also defined as the contribution weight of the $k^{\prime}$-th concept in the shallow layer to the $k$-th concept in the deep layer.
% a low semantic-level concept by shallow-layer units to the high semantic-level concept-harmonized unit in the deep layer.

Therefore, we can obtain the most critical shallow-layer concept to the target deep-layer concept by optimizing the following problem

\begin{equation} \label{Concept-harmonized hierarchical inference model one}
\begin{split}
& \arg\max_{k^{\prime}}   \nabla_{\mathbf{t}^{S,k^{\prime}}} f_{\mathbf{c}^{D,k}}(  \mathbf{A}^S) \\
% \approx  & \arg\max_{\mathbf{t}^{h-1,k^{\prime}}}  \mathbf{w}^{h,k}_{S} \cdot \mathbf{t}^{h-1,k^{\prime}}_{S} 
\end{split}
\raisetag{2\normalbaselineskip}
\end{equation}

It should be noticed that the optimal solution for the shallow-layer concept in Eq. (\ref{Concept-harmonized hierarchical inference model one}) is same with that in Eq. (\ref{Concept-harmonized hierarchical inference model}) when the sparsity of $\mathbf{\alpha}^{D,k}_{S}$ in Eq. (\ref{Concept-harmonized hierarchical inference model}) is set to 1.
It means that the optimization based on the concept directional-derivative can be considered as a special case of the concept-harmonized hierarchical inference model in Eq. (\ref{Concept-harmonized hierarchical inference model}).
%===

%=====
\subsubsection{\textbf{The instance-level CHAIN interpretation}}
%\textbf{The concept-harmonized hierarchical inference interpretation (CHAIN).}
For the interpretation of a specific net decision for a given input, the concept harmonizing model is optimized from the highest to the lowest semantic level. 
The net unit is harmonized with a concept in a similar semantic level.
Next, in the hierarchical inference model, we deduce the target net-output from the deepest layer to the shallowest layer.
Lastly, based on the above, the inference of net-units from the deep to the shallow layer can be represented as the inference of visual concepts from the high to the low semantic level.
Finally, we can obtain the concept-harmonized hierarchical inference for the interpretation of a specific net decision.
%According to concept-harmonized hierarchical inference, we can obtain the concept-tree for the interpretation of a specific net decision.
%It can also be viewed as the concept-harmonized hierarchical inference interpretation for the net decision.

\subsubsection{\textbf{The class-level CHAIN interpretation}}
For the interpretation of net decisions for images from a specific class, we build the class-level CHAIN by selecting the shared concepts among instance-level CHAIN for different instances in the same class.
The class-level concept contribution weight is the average of its instance-level weights in a class dataset. 
%In a class-based concept-tree, the concept inference weight of a low semantic-level concept to a high semantic-level concept is the sum of its instance-based concept inference weights in a class dataset. 
%In a class-based concept-tree, a high semantic-level concept is disassembled into low semantic-level concepts whose concept inference weights are computed by the sum of its instance-based concept inference weights in a class dataset. 
%The class-based CHAIN interpretation is learned from networks by utilizing a class dataset.
Therefore, it interprets the network mechanism from the class-level view.

\section{Experiments}
%In the experiment, we evaluate the CHAIN model on different networks and datasets.
In this section, the experiments show the qualitative and quantitative analyses for the performance of the proposed interpretation model.
In section \ref{instance}, we provide the instance-level CHAIN interpretation for the net being interpreted.
%In section \ref{instanceInter} and \ref{instanceIntra}, to explain the net prediction for an intance, inter-class and intra-class CHAIN interpretation.
In section \ref{class}, the CHAIN interpretation also applies to explain net predictions in class level. 
In section \ref{classIntra} and \ref{classInter}, networks can be further understood by its intra-class and inter-class CHAIN interpretation on the class level.

\subsection{Experimental setting}

\subsubsection{\textbf{ResNet on the Places365 scene classification dataset}}
In the experiment, CHAIN interpretation is applied to explain the ResNet-18 \cite{7780459} which is pretrained on the ImageNet dataset \cite{krizhevsky2012imagenet} and finetuned on the Places365 scene classification dataset \cite{7968387}.
ResNet is a convolutional neural network and can learn rich feature representations.
It can classify images into 365 scene categories. 
In the concept harmonizing model, we use five layers (i.e., the output, conv5, conv4, conv3, and conv2) to be harmonized with five semantic level concepts (scene, object, part, material, and color).

\subsubsection{\textbf{The concept harmonizing dataset}}
In the concept harmonizing model, the Broden Dataset is utilized as the concept dataset, which is a fully annotated image dataset \cite{8099837}.
The Broden dataset contains a hierarchical level of labeled visual-concept samples.
We use five semantic level concepts, i.e., the scene, object, part, material, and color concepts from the Broden dataset.
The annotations are mostly in pixel-level except for the scene annotation for image level.
The five semantic level concepts are from the ADE20K \cite{8100027}, Pascal-Part \cite{chen2014detect}, and OpenSurfaces datasets \cite{bell2014intrinsic}.

\subsubsection{\textbf{Inference distance}}
In the experiment, we define the inference distance to quantitatively analyze the CHAIN interpretation.

\textbf{Inference distance of the concept $c$ for the image set $s$} is defined as the average Euclidean distance of inference weights for the corresponding image set, which is calculated as
\begin{equation} \label{dist}
\begin{split}
& dist(concept_{c}, set_{s}) \\
&= \frac{1}{N_{i}} \sum_{i} \| \mathbf{w}^{concept_{c}}_{set_{s,i}}-\mathbf{\bar{w}}^{concept_{c}}_{set_{s}} \|_{2} \\
%&  \frac{1}{N_{i}} \sum_{i} \sqrt{(\mathbf{w}^{concept_{c}}_{set_{s,i}}-\mathbf{\bar{w}}^{concept_{c}}_{set_{s}})(\mathbf{w}^{concept_{c}}_{set_{s,i}}-\mathbf{\bar{w}}^{concept_{c}}_{set_{s}})^T}
\end{split}
\raisetag{2\normalbaselineskip}
\end{equation} 
where the center of inference distance is $\mathbf{\bar{w}}^{concept_{c}}_{set_{s}}$, which is calculated by
%The average Euclidean distances for the golden retriever, chow, and horse are 0.1544, 0.1259, and 0.0871, respectively.
\begin{equation} \label{dist}
\begin{split}
& \mathbf{\bar{w}}^{concept_{c}}_{set_{s}}  = \frac{1}{N_{i}} \sum_{i} \mathbf{w}^{concept_{c}}_{set_{s,i}}
\end{split}
\raisetag{2\normalbaselineskip}
\end{equation}

\textbf{Inference distance between the concept $c$ for the image set $s$ and the concept $c'$ for the image set $s'$} is obtained by
\begin{equation} \label{dist}
\begin{split}
& dist(concept_{c}, set_{s};concept_{c'}, set_{s'}) \\
&= \| \mathbf{\bar{w}}^{concept_{c}}_{set_{s}}-\mathbf{\bar{w}}^{concept_{c'}}_{set_{s'}} \|_{2} \\
%&  \frac{1}{N_{i}} \sum_{i} \sqrt{(\mathbf{w}^{concept_{c}}_{set_{s,i}}-\mathbf{\bar{w}}^{concept_{c}}_{set_{s}})(\mathbf{w}^{concept_{c}}_{set_{s,i}}-\mathbf{\bar{w}}^{concept_{c}}_{set_{s}})^T}
\end{split}
\raisetag{2\normalbaselineskip}
\end{equation}

\begin{figure*}[htp]
	\centering
	\includegraphics[width=1\linewidth]{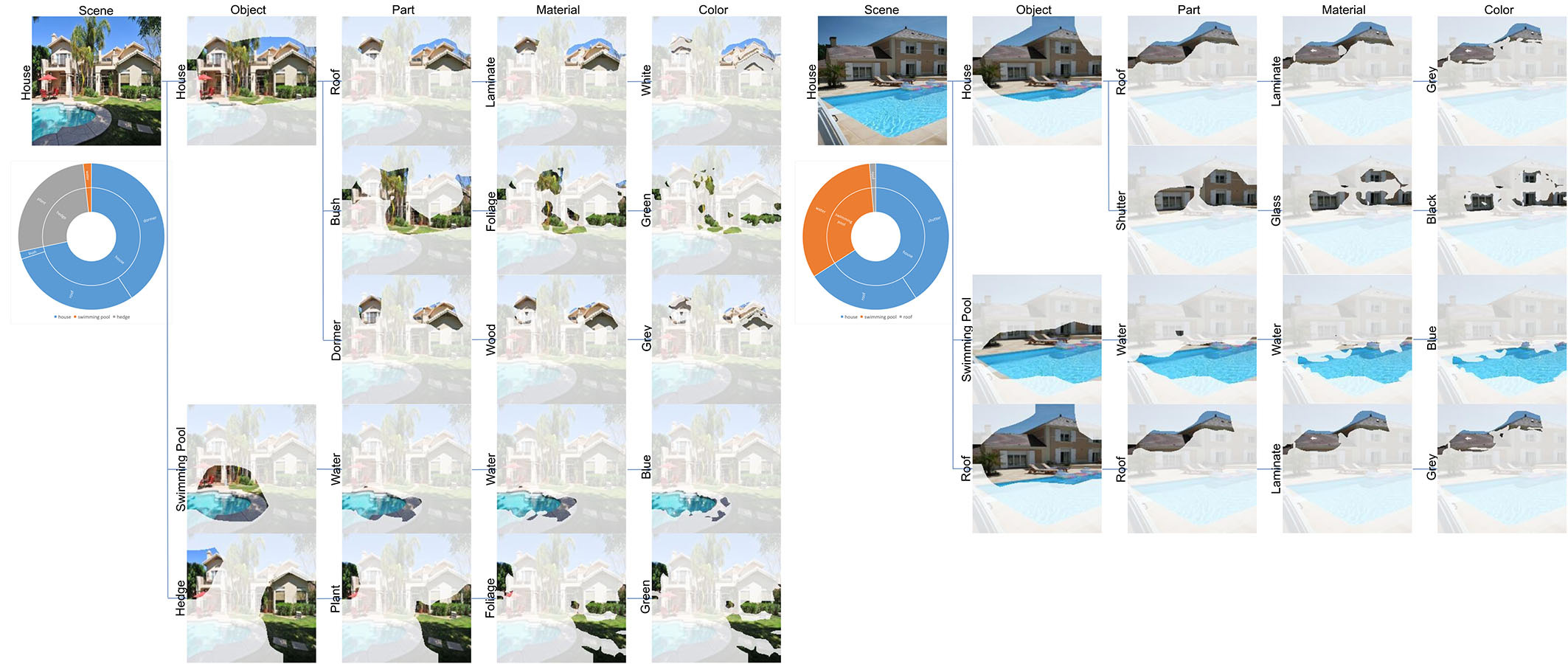}
	\centering
	\caption{Instance-level CHAIN interpretation for images of a house with a swimming pool.}
	\label{fig:house_instance_swim}
\end{figure*}

\begin{figure*}[htp]
	\centering
	\includegraphics[width=1\linewidth]{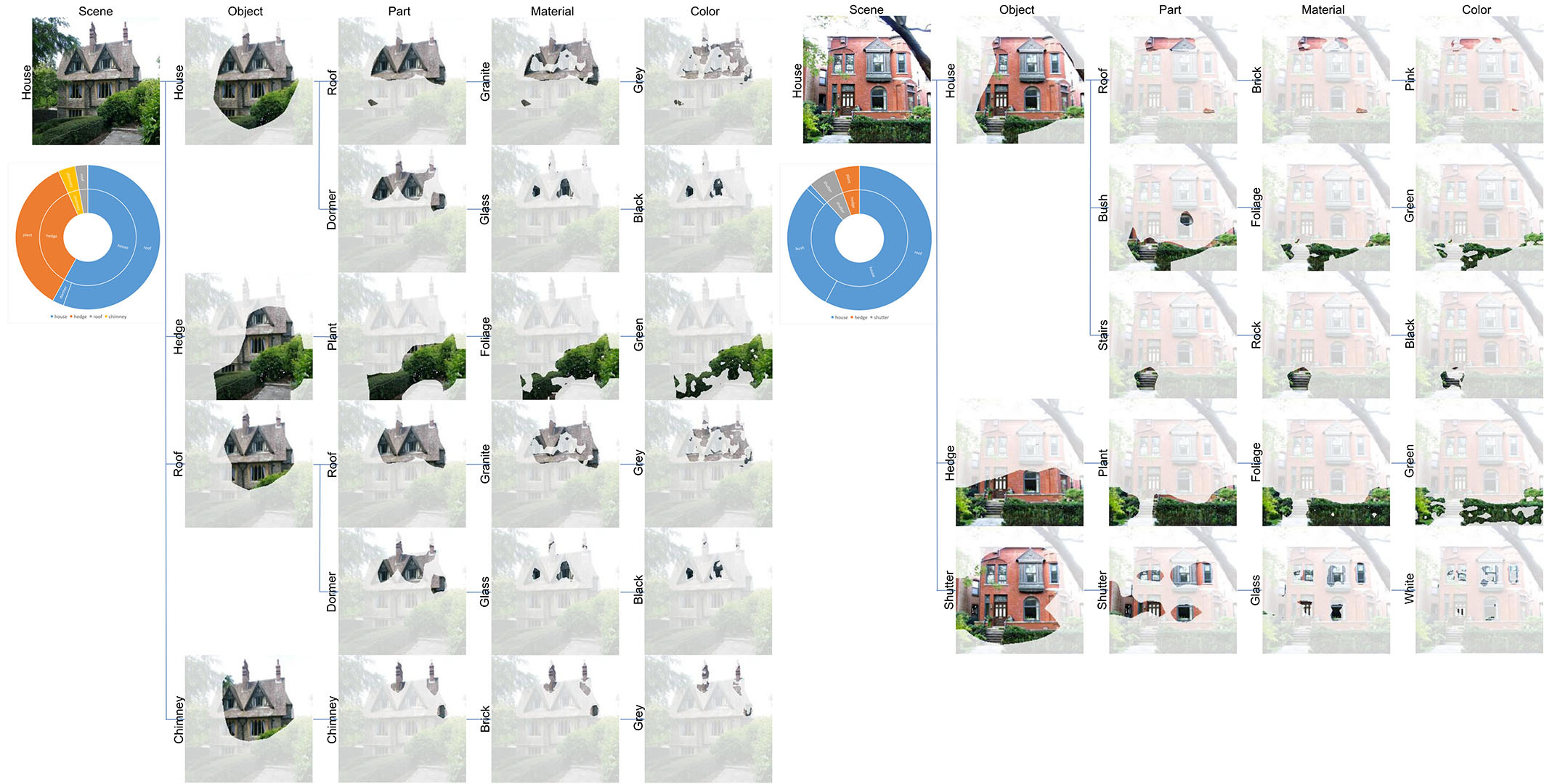}
	\centering
	\caption{Instance-level CHAIN interpretation for House-class images in which a house is enclosed the hedge.}
	\label{fig:house_instance_hedge}
\end{figure*}

\begin{figure*}[htp]
	\centering
	\includegraphics[width=1\linewidth]{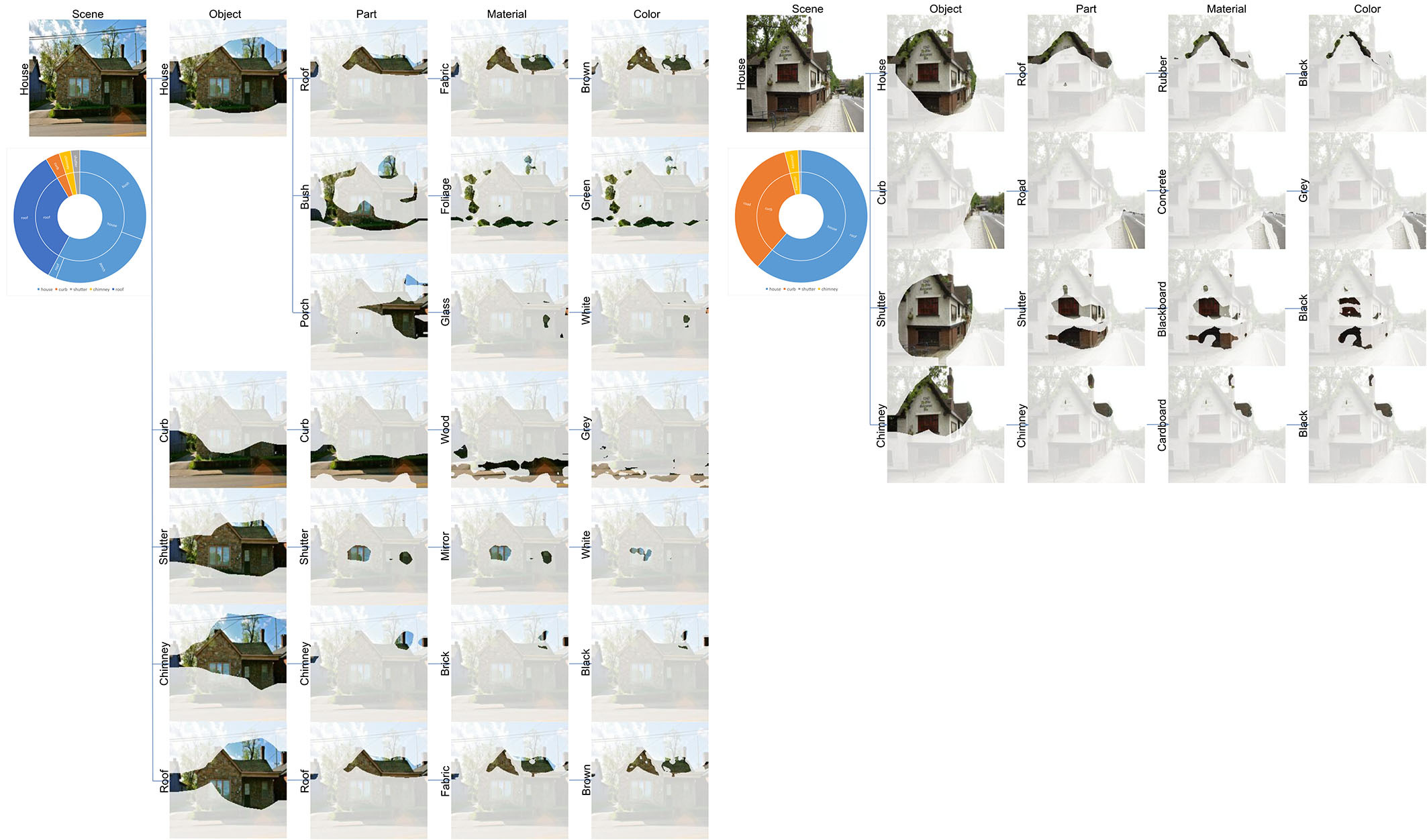}
	\centering
	\caption{Instance-level CHAIN interpretation for House-class images in which a house is by the roadside.}
	\label{fig:house_instance_curb}
\end{figure*}

\begin{figure*}[htp]
	\centering
	\includegraphics[width=1\linewidth]{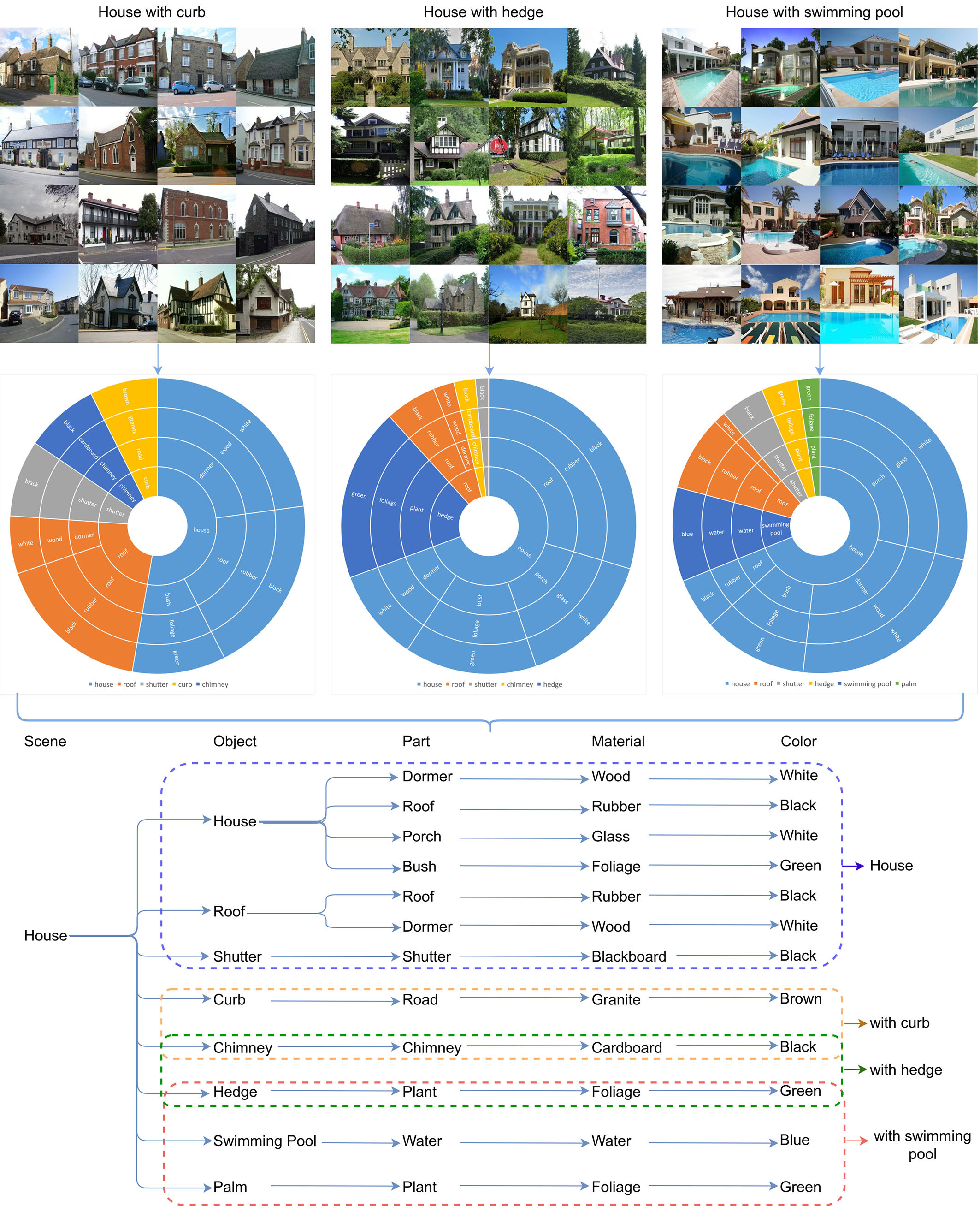}
	\centering
	\caption{Class-level CHAIN interpretation for the House class. (Top) House images with three types of surroundings, i.e. curb, hedge, and swimming pool. (Middle) Sunburst charts of images with different surroundings. The innermost ring of a sunburst chart shows concepts that are crucial to the House-class prediction. The expansion of a concept section to its outer ring shows the lower-level concepts that are important to the concept itself. (Bottom) CHAIN interpretation diagram for the House-class images. The fraction enclosed by the purple dashed line denotes the house-related concepts shared by the three types of images. Fractions in yellow, green, and red dashed rectangles are the concepts for different surroundings.}
	\label{fig:house_class}
\end{figure*}

\begin{figure*}[htp]
	\centering
	\includegraphics[width=1\linewidth]{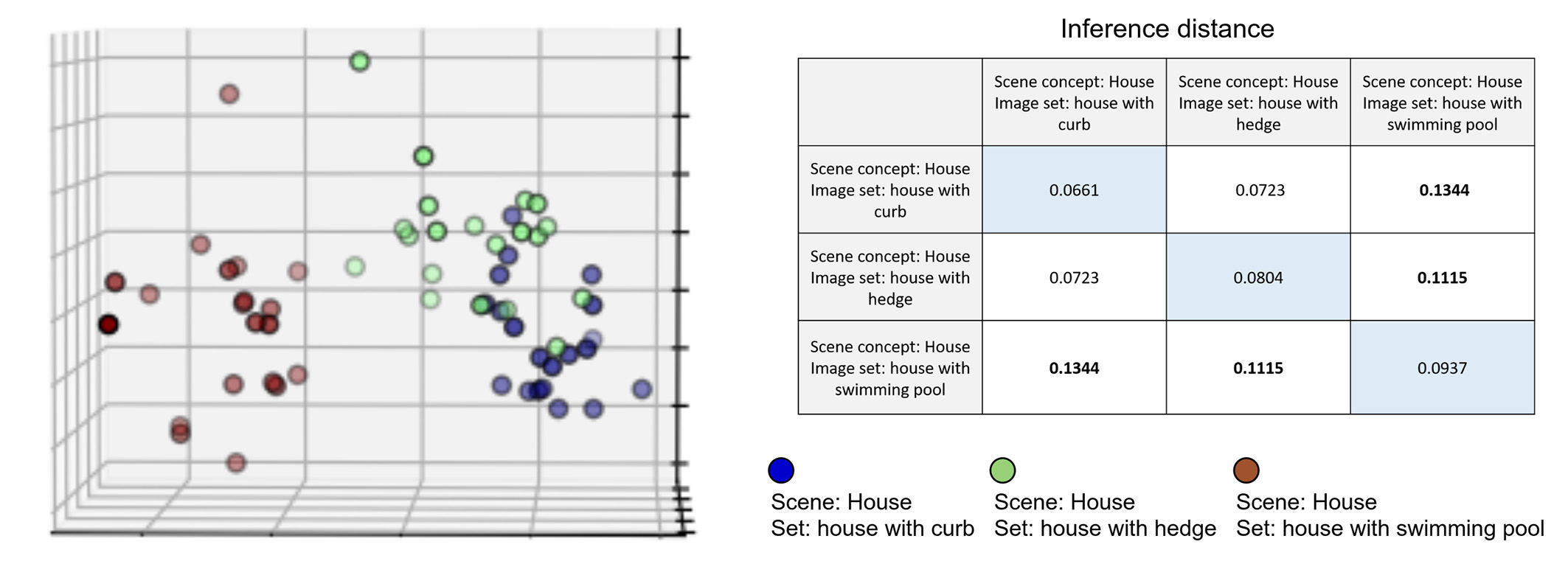}
	\centering
	\caption{Intra-class CHAIN interpretation for the House class on scene level. The left chart plots the inference weights of the house concept (scene level) for three image sets (houses with curb, hedge, and swimming pool) in the 3D-PCA space. The right table shows their inference distances.}
	\label{fig:intraShouse}
\end{figure*}

\begin{figure*}[htp]
	\centering
	\includegraphics[width=1\linewidth]{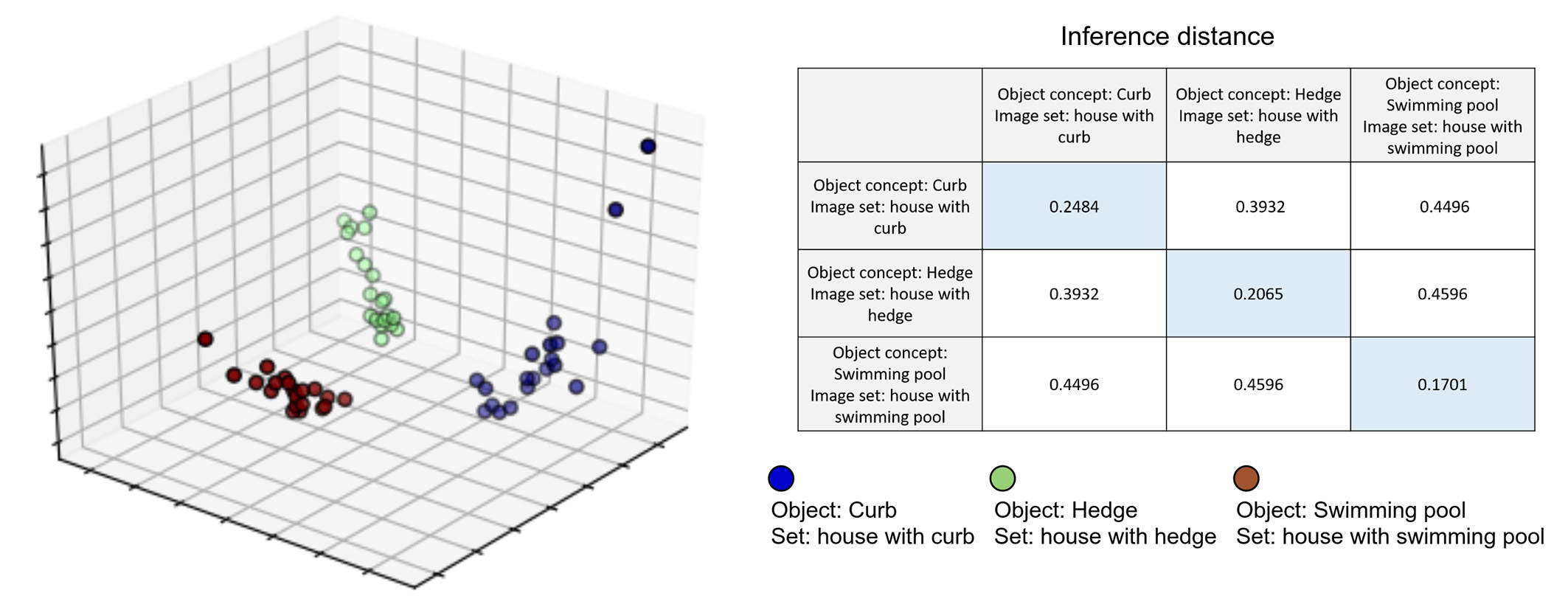}
	\centering
	\caption{Intra-class CHAIN interpretation for the House class on object level. The left chart plots the inference weights of three object concepts (curb, hedge, and swimming pool) for their corresponding image sets (houses with curb, hedge, and swimming pool) in the 3D-PCA space. The right table shows their inference distances.}
	\label{fig:intraOdiff}
\end{figure*}

\begin{figure*}[htp]
	\centering
	\includegraphics[width=1\linewidth]{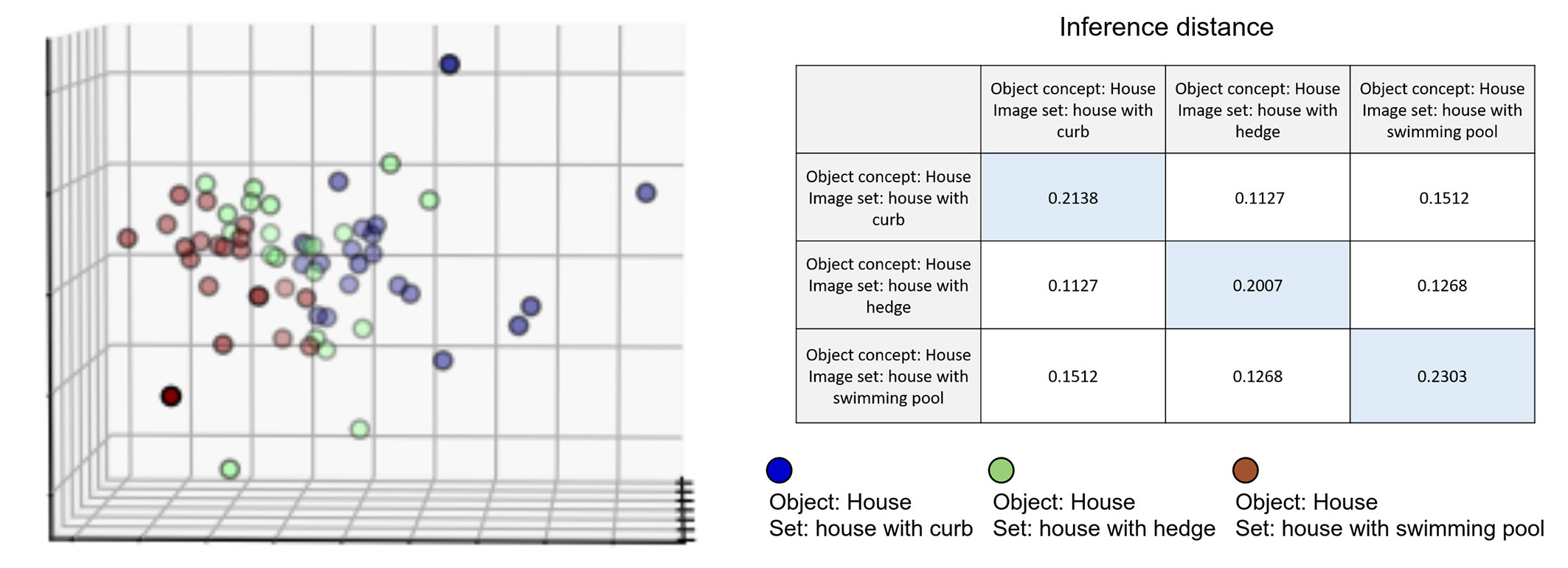}
	\centering
	\caption{Intra-class CHAIN interpretation for the House class on object level. The left chart plots the inference weights of the house concept (object level) for three image sets (houses with curb, hedge, and swimming pool) in the 3D-PCA space. The right table shows their inference distances.}
	\label{fig:intraOhouse}
\end{figure*}

\begin{figure*}[htp]
	\centering
	\includegraphics[width=1\linewidth]{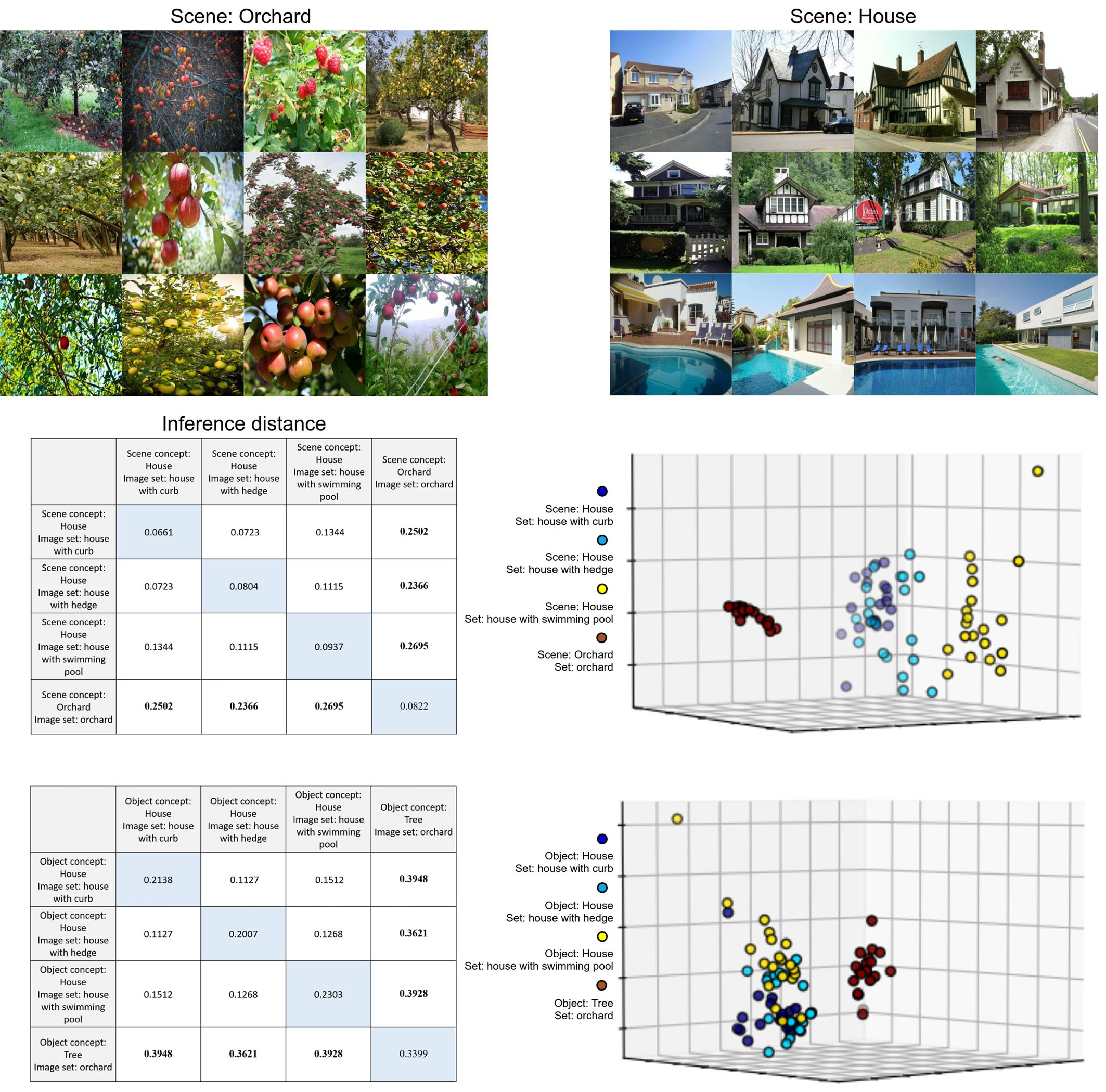}
	\centering
	\caption{Inter-class CHAIN interpretation for the Orchard and the House classes. The top row shows samples from four image sets (orchard, houses with curb, hedge, and swimming pool) for two classes. The right chart (in the middle) plots the inference weights of the scene concepts for four image sets in the 3D-PCA space. The left table (in the middle) shows their inference distances. Similarly, the bottom row shows the analysis for the object-level concepts.}
	\label{fig:inter}
\end{figure*}

\subsection{The instance-level CHAIN interpretation}\label{instance}

\subsubsection{The instance-level CHAIN interpretation for different classes}\label{instanceInter}

In this experiment, we randomly select four images from different classes to explain their net predictions.
For these images, the net accurately predicts their scene classes.

Fig. \ref{fig:instance-inter} shows the instance-level CHAIN interpretation for images from four classes (i.e. Farm, Orchard, House, and Pasture).
%For each image, the CHAIN model infers its net prediction from high semantic level to low semantic level.
For the interpretation of a specific net decision for a given input, CHAIN provides the concept-harmonized hierarchical inference for the network decision-making process from the highest to the lowest semantic level.
Meanwhile, the CHAIN interpretation provides visualization for concepts in each semantic level.

In the bottom right of Fig. \ref{fig:instance-inter}, for the pasture scene image, CHAIN infers that the pasture scene prediction is based on the house, grass, fence concepts which are learned from shallow layer features for object level.
Moreover, the horse concept in the object level is inferred from the torso concept at the part level which is deduced from the skin material concept.
Finally, the horse concept can be hierarchically deduced from the orange color concept.
The CHAIN interpretation is a logical decision-making process to explain the net decision.

Meanwhile, the visualization of concepts in the CHAIN interpretation can localize the corresponding visual parts, which can further interpret the net feature learning for visual concepts.
In the CHAIN interpretation, the net prediction is interpreted from the scene to the color semantic level.
Similarly, the scale of concept visualization is decreased from the high to the low semantic level.
The reason is that the receptive field in the net feature learning process is decreased from the deep to the shallow layer.

\subsubsection{The instance-level CHAIN interpretation within a class}\label{instanceIntra}
In this section, house images with three types of surroundings are selected as the target to analyze the CHAIN interpretation within a class.
Specifically, for each type, we randomly choose two instances to show their results.

Meanwhile, in the CHAIN interpretation for each instance, the sunburst chart presents object concepts (the inner circle) and their corresponding part concepts (the outer circle). 
The proportion of each visual concept in the inner circle indicates its contribution to the network scene prediction. 
Similarly,  the contribution of a part concept for its object concept is indicated by its proportion in the outer circle.

Fig. \ref{fig:house_instance_swim} shows instance-level CHAIN interpretation for House-class images in which a house is with a swimming pool.
In Fig. \ref{fig:house_instance_swim}, the CHAIN interpretation of these two images both includes house and swimming pool concepts in the object level, which is consistent with the visual perception for these images.
From visual understanding in the object level, the left image also contains a hedge region which does not exist in the right image.
In contrast, the house roof can be observed clearly in the right image compared with that in the left image.
These differences can be reflected in the corresponding CHAIN interpretation.
It means the CHAIN interpretation can explain the difference in net feature learning for different images in the set of the house with a swimming pool.

In Fig. \ref{fig:house_instance_swim}, the part level concept of the swimming pool is explained as water concept for both images.
In the left image, the part level concept for the house object concept includes the bush concept which is not deduced in the right image.

Fig. \ref{fig:house_instance_hedge} presents instance-level CHAIN interpretation for House-class images in which a house is enclosed the hedge.
For the image set of the house with the hedge, CHAIN can distill the difference and similarity in net feature learning as the concept interpretation for different images.
For example, the house and hedge object concepts are shared in CHAIN interpretation for both images in Fig. \ref{fig:house_instance_hedge}.
In contrast, roof and chimney object concepts only exist in the interpretation of the left image in which these two objects are apparent.

Fig. \ref{fig:house_instance_curb} shows instance-level CHAIN interpretation for House-class images in which a house is by the roadside.
From Fig. \ref{fig:house_instance_swim}, \ref{fig:house_instance_hedge} and \ref{fig:house_instance_curb}, it is noted that for the object concept level, the house object concept is the common interpretation for house images with different surroundings.
In comparison, the swimming pool, hedge, and curb object concepts are unique interpretations for corresponding house surroundings.
Therefore, at the instance-level, the CHAIN interpretation can interpret the difference and similarity of net feature learning within a class.

\subsection{The class-level CHAIN interpretation}\label{class}
In this section, we analyze the CHAIN interpretation on the class level.
Specifically, in section \ref{classIntra}, the house with three types of surroundings is selected for the intra-class analysis.
In section \ref{classInter}, scene orchard and house image sets are applied for the study of inter-class CHAIN interpretation.

\subsubsection{The intra-class CHAIN interpretation}\label{classIntra}
In this experiment, for the house scene class, the intra-class interpretation is analyzed by house images with three typical surroundings (i.e., curb, hedge, and swimming pool).
For each type of surrounding, we randomly choose twenty house images as the corresponding image set.

Fig. \ref{fig:house_class} displays class-level CHAIN interpretation for the House class. 
(Top) House images with three types of surroundings, i.e. curb, hedge, and swimming pool. 
(Middle) Sunburst charts of images with different surroundings. 
The innermost ring of a sunburst chart shows concepts that are crucial to the House-class prediction. 
The expansion of a concept section to its outer ring shows the lower-level concepts that are important to the concept itself.
Meanwhile, in each sunburst chart, the concepts in each level are sorted in descending order of the contribution to their corresponding high level concept.
(Bottom) CHAIN interpretation diagram for the House-class images. 
The fraction enclosed by the purple dashed line denotes the house-related concepts shared by the three types of images. 
Fractions in yellow, green, and red dashed rectangles are the concepts for different surroundings.

In the three sunburst charts of Fig. \ref{fig:house_class}, on the object level, the house concept has the most significant contribution to the house scene prediction.
Meanwhile, hedge and swimming pool object concepts own the second-largest contribution to the net prediction on house image sets for hedge and swimming pool, respectively.
For the house image set with the curb, the object concept curb also contributes a lot to the house prediction.
Therefore, CHAIN can explain the intra-class net predictions by presenting their common and unique concepts within a class.

In the bottom of Fig. \ref{fig:house_class}, we present the CHAIN interpretation for the house class in which the net output is inferred from the scene to the color semantic level.
The class level CHAIN interpretation (in the bottom of Fig. \ref{fig:house_class}) is consistent with the observation of the three sunburst charts.
In the class level CHAIN interpretation, the hedge object concept is deduced to plant part, and then to the foliage material, and finally to the green color concept.
Meanwhile, the hedge object concept is shared in the interpretation of the house with the hedge and swimming pool.
From the image level, it is observed that many images in the swimming pool set involve hedge region, as shown in the second image in the first row and the fourth image in the third row.
Therefore, the observation of CHAIN interpretation is understandable.
Similarly, the chimney as the common object concept in the CHAIN interpretation can be found in both image sets for curb and hedge.

Fig. \ref{fig:intraShouse} depicts the intra-class CHAIN interpretation for the House class at the scene level. 
The left plot shows the inference weights of the house concept (scene level) for three image sets (houses with curb, hedge, and swimming pool) in the 3D-PCA space. 
The right table shows their inference distances.

From the table in Fig. \ref{fig:intraShouse}, the numbers in the bold font (0.1344 and 0.1115) are larger than the others.
From the left plot, the red points can be easily separated from the other points.
In comparison, the green and blue points have some overlap.
Therefore, in the CHAIN interpretation, at the scene level, the inference of the house concept for the swimming pool set is different from those for the curb and hedge sets.
%From the visual perception, the house type for the swimming pool set is more luxurious than those for the curb and hedge sets.
From the visual perception of the scene, the house surroundings for the swimming pool set has a huge difference from those for the curb and hedge sets.

Fig. \ref{fig:intraOdiff} presents the intra-class CHAIN interpretation for the House class on object level. The left chart plots the inference weights of three object concepts (curb, hedge, and swimming pool) for their corresponding image sets (houses with curb, hedge, and swimming pool) in the 3D-PCA space. The right table shows their inference distances.

In the left plot of Fig. \ref{fig:intraOdiff}, the three color points can be clustered into three groups separately.
In the right table, diagonal entries are smaller than the others.
Hence, at the object level of the CHAIN interpretation, the inference for three object concepts (curb, hedge, and swimming pool) can be easily distinguished between each other.

Fig. \ref{fig:intraOhouse} shows the intra-class CHAIN interpretation for the House class on object level. The left chart plots the inference weights of the house concept (object level) for three image sets (houses with curb, hedge, and swimming pool) in the 3D-PCA space. The right table shows their inference distances.

The left plot in Fig. \ref{fig:intraOhouse} shows that the three color points mix with each other, which is also testified by the right table.
It means at the object level, the inference of the house concept for three image sets are similar.
For the three image sets, the house object is similar even though the difference of their surroundings.
In summary, the CHAIN interpretation can learn the similarity and variance within a class, which is consistent with our visual understanding.

\subsubsection{The inter-class CHAIN interpretation}\label{classInter}
In this experiment, the scene orchard and house are utilized for the study of the intra-class CHAIN interpretation.
For the orchard class, we randomly select twenty images for testing.
For the house class, we continue using the previous three house image sets.

Fig. \ref{fig:inter} displays the inter-class CHAIN interpretation for the Orchard and the House classes. The top row shows samples from four image sets (orchard, houses with curb, hedge, and swimming pool) for two classes. The right chart (in the middle) plots the inference weights of the scene concepts for four image sets in the 3D-PCA space. The left table (in the middle) shows their inference distances. Similarly, the bottom row shows the analysis for the object-level concepts.

In the left two tables of Fig. \ref{fig:inter}, the entries in bold font are more significant than the others. 
In the right two plots, the data in red color can be easily separated from the other data.
At the scene level, the CHAIN interpretation of the scene concept orchard varies a lot from that of the scene house.
Likewise, there exists a large discrepancy between the interpretation of the house object concept and that of the tree object concept.
Therefore, the inter-class difference between orchard and house is larger than the intra-class difference, which is also aligned with the visual perception from images.
The CHAIN interpretation can be used for the inter-class investigation.

\section{Conclusion} 
\label{sec:conclusion}
In this paper, the CHAIN interpretation is proposed to give an explanation for the net decision-making process. 
Specifically, the CHAIN interpretation hierarchically reasons a net decision to be visual concepts from the high level to the low level.
The hierarchical visual concepts also help explain the layer structure of the network.
Except for the instance-level interpretation, the CHAIN interpretation can also provide inference at the class level.
Experiment results demonstrate that the proposed CHAIN model can provide reasonable interpretations at both levels.
%In the work, we proposed a novel CHIP model, which can provide visual interpretation for the predictions of networks without requiring the retraining of networks.
%Further, we combine the visual interpretation in the first and the last convolutional layers to obtain Refined CHIP visual interpretation that is class-discriminative and high-resolution.
%Through experiment evaluation, we have demonstrated that the proposed interpretation model can provide more reasonable visual interpretation compared with previous methods. 
%The proposed method can also outperform other visual interpretation methods in the application of weakly-supervised object localization in ILSVRC 2015 benchmark.
%
\bibliographystyle{IEEEtran}
\bibliography{chain_ref}

\end{document}